\documentclass[3p,11pt]{elsarticle}

\usepackage{lineno,hyperref}
\modulolinenumbers[5]

\usepackage{tikz}
\usetikzlibrary{decorations.pathreplacing}

\usepackage{graphicx}
\usepackage{amsmath,amssymb}
\usepackage{algorithmic}
\usepackage{algorithm, url, pifont, multirow}
\usepackage{caption}
\usepackage{subcaption}
\usepackage{todonotes} 
\usepackage{xcolor}
	
\newcommand{\reels}{\mathbb{R}}
\def\btheta{\boldsymbol{\theta}}
\def\bmu{{\boldsymbol{\mu}}}
\def\bSigma{{\boldsymbol{\Sigma}}}

\newtheorem{Rem}{Remark}

\journal{Mechanical Systems and Signal Processing}

\newcommand{\parpapb}[4]{\frac{\partial {#1}_{#2}}{\partial {#3}_{#4}}}

\newcommand{\bx}{\boldsymbol{x}}
\newcommand{\bX}{\boldsymbol{X}}
\newcommand{\bomega}{\boldsymbol{\omega}}
\newcommand{\esp}{\mathbb{E}}

\begin{document}

\sloppy 

\begin{frontmatter}

\title{Clustering acoustic emission data streams with sequentially appearing clusters using mixture models}

\author{Emmanuel Ramasso}
\address{Institut FEMTO-ST (UMR CNRS 6174), Universit\'e Bourgogne Franche-Comt\'e, Département Mécanique Appliquée, Besan\c con, France}
\ead{emmanuel.ramasso@femto-st.fr}
\ead[url]{https://github.com/emmanuelramasso}

\author{Thierry Den\oe ux}
\address{Universit\'e de technologie de Compi\`egne, CNRS, Heudiasyc, Compi\`egne, France \\
Institut universitaire de France, Paris, France}
\ead{thierry.denoeux@utc.fr}
\ead[url]{https://www.hds.utc.fr/~tdenoeux/}

\author{Ga\"el Chevallier}
\address{Institut FEMTO-ST (UMR CNRS 6174), Universit\'e Bourgogne Franche-Comt\'e, Département Mécanique Appliquée, Besan\c con, France}
\ead{gael.chevallier@univ-fcomte.fr}

\begin{abstract}
The interpretation of unlabeled acoustic emission (AE) data classically relies on general-purpose clustering methods. While several criteria have been used in the past to select the hyperparameters of those algorithms, few studies have paid attention to the development of dedicated objective functions in clustering methods able to cope with the specificities of AE data. We investigate how to explicitly represent clusters onsets in mixture models in general, and in Gaussian Mixture Models (GMM) in particular. We propose the first clustering method able to provide, through parameters estimated by an expectation-maximization  procedure, information about when clusters occur (onsets), how they grow (kinetics) and their level of activation through time. This new objective function accommodates continuous timestamps of AE signals and, thus, their order of occurrence. The method, called \textsf{GMMSEQ}, is experimentally validated to characterize the loosening phenomenon in bolted structure under vibrations. A comparison with four standard clustering methods on  raw streaming data from five experimental campaigns  shows that \textsf{GMMSEQ} not only provides useful qualitative information about the timeline of clusters, but also shows better performance in terms of cluster characterization. 
\end{abstract}


\end{frontmatter}


\section{Introduction}
\label{intro}


ASTM standard E1316 \cite{ASTME1316,ABS239} defines Acoustic Emission (AE) 
as the detection of the subnanometric displacements of the surface of a material induced by the propagation of an elastic wave generated by a sudden and permanent change in the material integrity. This capability makes the AE technique particularly relevant to gain insights into the behavior of a material, a structure or an equipment under usage  \cite{scruby1987introduction,FarrarBook,awerbuch2016applicability,bhuiyan2018toward,he2021overview,alshorman2021sounds} and accounts for its wide use in applications related to material testing, Structural Health Monitoring (SHM) and process monitoring and control.  

\begin{figure}
\centering
\includegraphics[width=0.99\textwidth]{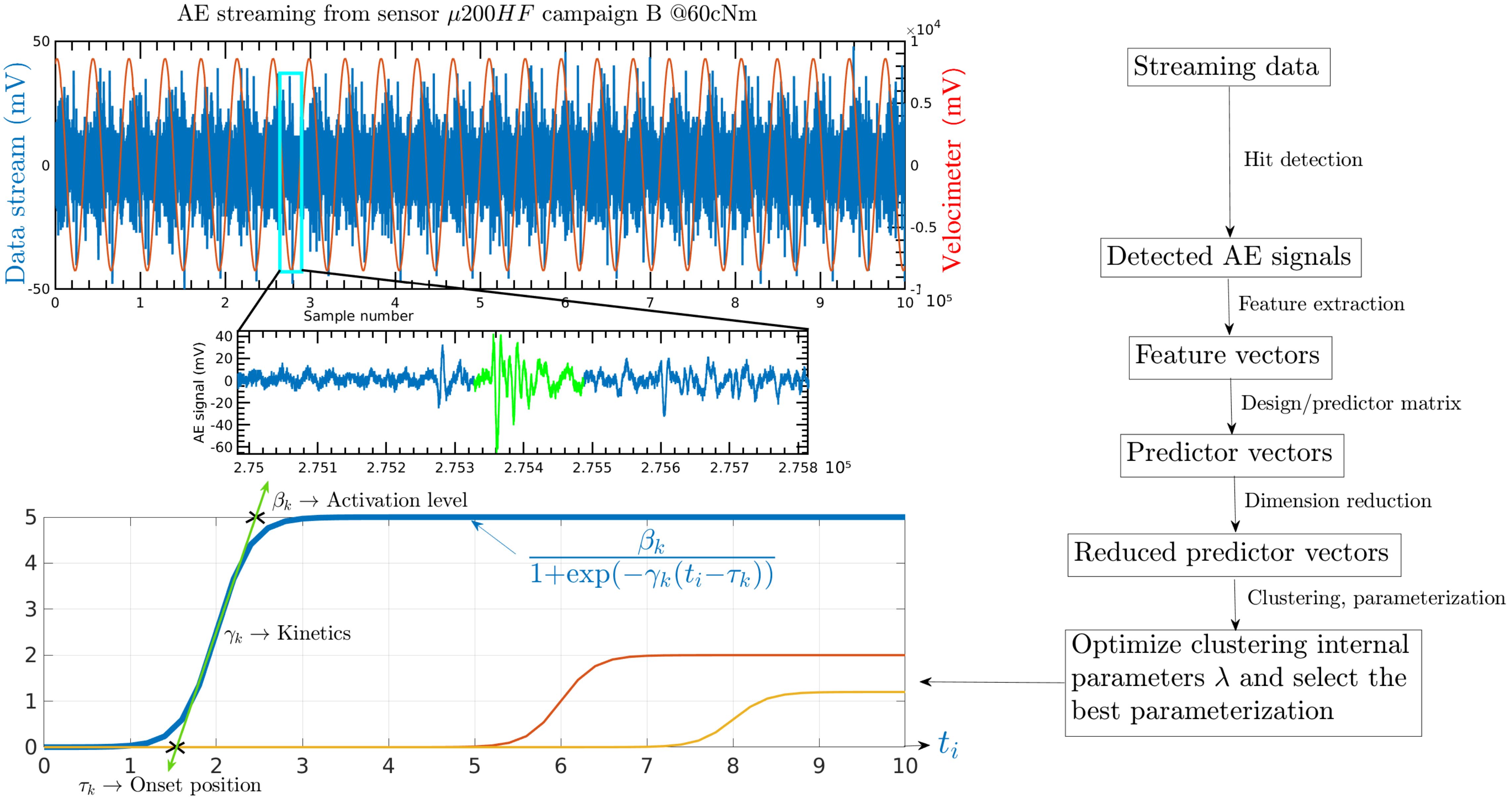}
\caption{Proposed methodology  and illustration on a real data stream extracted from one of the campaigns. \label{zdzzdzefzfz}}
\end{figure}

Original AE data take the form of a \emph{data stream} recorded by sensors attached onto a structure (Figure  \ref{zdzzdzefzfz}). The sensors, converting the subnanometric displacements into voltage signals, have to be read continuously in order to catch all events originating from the material. 
The data stream is then segmented using a wave-picking algorithm with the aim to detect damage-related (non-noise) AE signals \cite{kurz2005strategies,Pomponi2015110,bianchi2015wavelet,WARRENLIAO201074,kharrat2016signal,madarshahian2019acoustic}. In feature-based interpretation of AE signals, a \emph{feature extraction} step is performed in which  AE signals are represented in a common feature space. The set of feature vectors represents an AE {{data set}} for a given experiment and is generally stored in an $N\times d$ feature matrix 
\begin{equation}
\bX = [\bx_1^\intercal, \dots,\bx_i^\intercal,\dots, \bx_N^\intercal]
\end{equation}
where $\bx_i^\intercal \in \reels^d$ is the transposed feature vector computed from the $i$-th AE signal.
The timestamps of AE signals are their instants of occurrence  and are denoted as $t_i$ with
\begin{equation}
0=t_0 < t_1< \dots<t_i< \dots< t_N=T, 
\end{equation}
where $T$ is the data stream duration. AE data have several special characteristics in terms of data processing \cite{RAMASSO2020103478}; in particular timestamps are continuous and \emph{unequally-spaced in time}, i.e., we generally have
\begin{equation}
\frac{t_i - t_{i-1}}{t_j - t_{j-1}} \neq 1. \label{mljkdpozdzdz}
\end{equation}
For a given AE signal, the $d$ features generally belong to a standard list of AE features, some of which are listed, for example, in \cite{Kattis17}. In the sequel, we suppose that the features have been extracted using the algorithm introduced in \cite{kharrat2016signal}, which implements in MATLAB\textsuperscript{\tiny\textregistered} common features available in the Mistras AEWin\textsuperscript{\tiny\textregistered} software. Unsupervised feature selection is not tackled in this paper. The reader interested in this topics can refer, for instance, to the aforementioned references and \cite{manson2001visualisation,li2012feature,doan2015unsupervised,sause16}.

Since a huge amount\footnote{The number of AE signals depends on the materials and the type of loading. For example,  hundreds of  thousands of  signals were detected for composite materials during quasi-static tests reported in \cite{ramasso2015unsupervised}.} of AE signals can be detected in a data stream, it is  difficult, except for specific configurations, to know the ground truth for a sufficient large amount of AE signals. Lack of knowledge about the source of AE signals prevents us from using supervised learning methods for AE data interpretation, or even for anomaly detection when ``normal'' condition data are available \cite{MARTINDELCAMPO2017187,FUENTES2020776}. \emph{Unsupervised learning}, particularly based on clustering, is generally used to extract information from AE data. This is the main scope of the present work.
  
\emph{Clustering} methods have been applied for decades to interpret AE data \cite{wang2019tool,chelliah2019optimization,zhou2018cluster}. The most commonly used methods are the K-means \cite{MacQueen1967,chai2017acoustic}, the fuzzy C-means (FCM) \cite{dunn1973fuzzy,1198989}, the Gustafson-Kessel (GK) algorithm \cite{gustafson1979fuzzy,ramasso2015unsupervised} and Gaussian Mixture Models (GMM) \cite{mclachlan1988mixture,sawan2015unsupervised}. 
A clustering method computes  membership degrees of  feature vectors to  clusters; a hard partition is then obtained by assigning each vector to its maximum-membership cluster. Based on this partition, data interpretation is generally necessary to determine the correspondence between clusters and damages. 
From a statistical point of view, the criteria used in  classical clustering methods rely on the assumption that AE signals are \emph{independent and identically distributed} (iid). Therefore, the partition, after applying one of those methods, does not depend on the ordering. From a physical point of view, this seems counterintuitive in most of AE-related applications since the progression of a damage type is known to depend on the preceding damage states and damage accumulation \cite{kaminski2015fatigue,saxena2011accelerated}. To cope with this problem, we need a clustering method taking the time distribution of AE signals into account. Note that ``time'' can be replaced by any monotonically increasing measure such as cumulative loading or cycles. 

Clustering methods dedicated to time-series have been developed in the past \cite{PATTERNRECONliao2005clustering,fu2011review} but only a few of them are able to manage continuous and uneven timestamps as shown in the recent review \cite{belhadi2020space}. One of the first attempts to modify standard approaches (such as K-means, FCM or GMM) so as to accommodate temporal data with such timestamps was presented in \cite{moller2003fuzzy}. The authors modeled the time series as piece-wise linear functions and proposed a distance measure between slopes. Using this distance, they derived a modified version of the FCM method. 

Clustering methods describe a data set through a set of parameters such as, for example, the cluster centers in the K-means algorithm or the means, covariance matrices and proportions  in Gaussian Mixture Models. Parameters are identified from data by iteratively optimizing an objective function that is an explicit function of these parameters. Additionally, clustering methods also depend on hyperparameters such as, e.g. the number of clusters, which cannot be optimized in this way. For hyperparameters, we need to perform a grid search by varying them and evaluate their impact on the clustering result. We also need a selection criterion (for determining, e.g., the optimal number of clusters).  For each set of  hyperparameter values, the objective function is first optimized to estimate the parameters of the clustering method. Then, the best configuration is selected. The learning process may, thus, be time-consuming. 

In AE,  hyperparameters are determined, in the majority of cases,  by computing a criterion such as Davies-Bouldin or Silhouette,  which focus on the shape of clusters. As a consequence, to be selected, hyperparameters must lead to compact and well-separated clusters in the feature space \cite{sawan2015unsupervised,sause12,Godin18,monti2016mechanical,SHIRAIWA20202791}. This is the definition of natural clusters often used in AE, originally based on \cite{carmichael1968finding}.  An alternative is to find hyperparameters that lead to clusters characterized by onsets which are well distributed in time or load. This approach was developed recently because some authors found limitations in shape-based criteria for interpreting AE data \cite{placet2013online,ramasso2015unsupervised,hohl2018computationally,rastegaev2018using,8529965,Vinogradov20, RAMASSO2020103478,NehaPhD}. 
However, to the best of our knowledge, \emph{the onset times of clusters have not been considered in an objective function so far}. {{\it Taking into account  onset times for clustering  AE data is the main objective of the present work. The proposed clustering method, called \textsf{GMMSEQ}, treats onsets as parameters that can be optimized together with the other cluster parameters directly from data, which was not possible before.}}

\textsf{GMMSEQ} relies on a modification of the original GMM to account for the fact that AE signals are indexed  by continuous timestamps.  More specifically, the proportions in the {{mixture}} are assumed to vary in time according to a model of evolution based on  sigmoid functions (Figure \ref{zdzzdzefzfz}). Each sigmoid function allows us to represent:
\begin{itemize}
\item The level of activation of a given damage  related to the cumulative number of signals generated by this damage;
\item The growth rate of the damage driven by the slope of the sigmoid function at the origin  and related to the kinetics of the damage; 
\item The instant of the damage onset. 
\end{itemize}
Therefore, this new clustering method makes it possible to identify when a damage first occurs (onset), how it grows (kinetics) and how it accumulates (cluster progression). {{The method makes a step beyond the standard approach to AE analysis by characterizing  damage progression through three parameters estimated from the data.}}

The ability to represent onset times, kinetics and activation level makes this approach relevant for applications in which the chronology, sequence or timeline are of key importance. {{For example, Sawan et al. argued for {\it an AE analysis approach that seeks to separate observations into the greatest number of clusters with distinct evolution behavior}. For that, they used a GMM in its original form and represented their results by means of a cumulated number of hits per cluster. However, the analysis is subjective without a proper identification of parameters related to the evolution, which is a common problem in AE analysis based on damage progression. In a previous paper \cite{ramasso2015unsupervised}, a criterion was proposed to identify clusters with different proportions and the authors experimentally observed that the timeline suggested by the clusters (obtained by different methods) was improved compared to standard shape-based criteria.}}

Another advantage of \textsf{GMMSEQ} is that the compactness and separability of clusters in the feature space is still of primary importance because it is managed explicitly using a mixture model with the possibility to adapt the distribution  to the data. While being developed to take AE data characteristics into account, {{this new clustering method can be applied to other temporal data for which onsets, growths and cumulative progression of clusters are relevant to the analysis.}} The proposed optimization procedure assumes that all data are available at once (offline analysis). 

The model and the estimation algorithm  are presented in Section \ref{sec:1}. The method is then illustrated in Section~\ref{sec:3} using simulated and real {{data sets}}. {{Data sets}} and codes are shared on Dataverse \cite{orionaedata} and Github\footnote{The project is publicly available at \url{https://github.com/emmanuelramasso/MIXMOD_SEQUENTIAL}.}.

\section{Gaussian Mixture Model with sequentially appearing clusters (\textsf{GMMSEQ})}
\label{sec:1}

The \textsf{GMMSEQ} method introduced in this paper is based on a GMM with time-varying proportions. The model is described in Section \ref{subsec:model}, and parameter estimation is addressed in Section \ref{subsec:estimation}. 

\subsection{Model}
\label{subsec:model}

\paragraph{Gaussian mixture models} In a mixture model, the data are supposed to follow a probability distribution defined as a weighted sum of $K$ distributions:
\begin{equation}
p(\bx_i ;\boldsymbol{\theta}) = \sum_{k=1}^K \pi_k g(\bx_i;\boldsymbol{\theta}),
\label{eq:MM}
\end{equation}
where $\pi_k$ denotes the proportion of component $k$ and $g$ can be, for example, a Gaussian, Gamma or Student-t probability density function (pdf); the vector of all parameters is represented by $\boldsymbol{\theta}$. In AE data clustering, GMM's  \cite{mclachlan1988mixture} have been widely used \cite{sawan2015unsupervised,FRIGIERI2016230,PREM201728,SAGAR2018647,DAS201942,FUENTES2020776} and are considered in the following developments. Each component in the mixture \eqref{eq:MM} is then a Gaussian pdf:
\begin{equation}
\phi(\bx_i;\boldsymbol{\mu}_k,\boldsymbol{\Sigma}_k)=\frac{1}{\sqrt{(2\pi)^d|\boldsymbol{\Sigma}_k|}}
\exp\left(-\frac{1}{2}(\bx_i-\boldsymbol{\mu}_k)^T{\boldsymbol{\Sigma}_k}^{-1}(\bx_i-\boldsymbol{\mu}_k)
\right),
\label{eq:gausspdf}
\end{equation}
where $\boldsymbol{\mu}_k$ and $\boldsymbol{\Sigma}_k$ are, respectively, the mean and covariance matrix of component $k$. 

After observing a realization  $\bx_1,\ldots,\bx_N$ from an  iid sample, the likelihood function is
\begin{equation}
L(\btheta;\bx_1,\ldots,\bx_N) = \prod_{i=1}^N \sum_{k=1}^K \pi_k \phi(\bx_i;\bmu_k,\bSigma_k),
\label{eq:GMM1}
\end{equation}
where $\btheta=(\bmu_1,\ldots,\bmu_K,\bSigma_1,\ldots,\bSigma_K,\pi_1,\ldots,\pi_{K-1})$. The maximum likelihood estimates (MLE's) cannot be computed in closed form and are usually computed numerically using the Expectation-Maximization (EM) algorithm \cite{dempster77}. 


\paragraph{New model} We propose to modify  \eqref{eq:GMM1} in order to incorporate a time-dependency of the data through the proportions:
\begin{equation}
p(\bx_1,\ldots,\bx_N;\btheta) = \prod_{i=1}^N \sum_{k=1}^K \pi_{\mathbf{i}k} \phi(\bx_i;\bmu_k,\bSigma_k),
\label{eq:GMM2}
\end{equation}
where the bold subscript in $\pi_{\mathbf{i}k}$ emphasizes the difference with  \eqref{eq:GMM1}, i.e., the proportion of each cluster $k$ are now dependent on the timestamps $t_i$  through additional variables $\alpha_{ik}$: 
\begin{equation}
\label{eq:props}
\pi_{ik} = \frac{\alpha_{ik}}{\sum_{\ell=1}^K \alpha_{i\ell}}, \quad k=1,\dots, K ,
\end{equation}
where $\alpha_{i1}=1$ for $i=1,\ldots,N$ and
\begin{equation}
\alpha_{ik} = \frac{\beta_{k}}{1+\exp[-\gamma_k(t_i-\tau_{k})]}, \quad k=2,\dots, K, \quad i=1,\ldots,N.
\label{eq:alpha}
\end{equation}
Parameters $\tau_k$, $\beta_k$ and $\gamma_k$ in the logistic (sigmoid) activation functions  \eqref{eq:alpha} must satisfy the following constraints:
\begin{equation}
\label{eq:constr}
0\le \tau_k \le T, \quad \beta_k \ge 0, \quad \gamma_k \ge 0
\end{equation}
for $k=2,\ldots,K$. As illustrated in Figure \ref{zdzzdzefzfz} (bottom-left), the degree of activation $\alpha_{ik}$ of the $k$-th cluster depends on the real timestamps of AE signals through a sigmoid function delayed by $\tau_{k}$, with upper limit $\beta_k$ and slope $\gamma_k$. The delay $\tau_{k}$ represents the onset time of cluster $k$. The proportions $\pi_{ik}$ in \eqref{eq:props} are equal to the normalized activation degrees. An example of how proportions can vary in time is shown in Figure \ref{fig:ffezlkjfekzlejfzfzfzf}b.

{{

The apparently simple modification of  GMM brought about by making the proportions in \eqref{eq:GMM1} dependent on time allows us to tackle the problem mentioned in Section \ref{sec:1}, concerning the inability of standard approaches to manage continuous and irregularly spaced-in-time timestamps. By associating a  sigmoid function to each cluster, \textsf{GMMSEQ} is, to our knowledge, the first clustering method dedicated to AE data able to estimate, directly from data, parameters related onsets, growth and kinetics of clusters. Its performance will be studied in Section \ref{sec:3}. Parameter estimation in this new model  requires a specific, and more complex EM algorithm, described in the next section.

}}

\subsection{Parameter estimation}
\label{subsec:estimation}

Maximum-likelihood parameter estimation in GMM's is usually carried out using the EM algorithm \cite{mclachlan08}, an approach that will also be  used here. However, {{there is no closed-form expression to update  parameters $\tau_k$, $\beta_k$ and $\gamma_k$ in the M-step, which makes it necessary to use a gradient algorithm. The algorithm is described in detail below.} }

\paragraph{Objective function}

The first step is to write down the complete-data log-likelihood function for our model{{:}}
\begin{equation}
    \ell_c(\btheta) = \sum_{i=1}^N\sum_{k=1}^K y_{ik} \log \pi_{ik} + y_{ik}\log \phi(\bx_i;\boldsymbol{\mu}_k,\boldsymbol{\Sigma}_k), 
\end{equation}
where $\btheta = (\{\bmu_k, \bSigma_k\}_{k=1}^K, \{\tau_k, \beta_k, \gamma_k\}_{k=2}^K)$ is the parameter vector, and the $y_{ik}$'s are binary cluster-membership indicator variables such that $y_{ik}=1$ if  observation $i$ belongs to $k$, and $y_{ik}=0$ otherwise. Here,  variables $y_{ik}$ are missing. At each iteration $q$ of the EM algorithm, we thus replace $\ell_c$ by its conditional expectation given the observed data, which yields the so-called \textit{auxiliary function} $Q$  \cite{dempster77}:
\begin{equation}
\label{eq:Q}
    Q(\btheta,\btheta^{(q)}) = \underbrace{\sum_{i=1}^N\sum_{k=1}^K  y_{ik}^{(q)} \log \pi_{ik}}_{Q_1} + \underbrace{\sum_{i=1}^N\sum_{k=1}^K y_{ik}^{(q)} \log \phi(\pmb{x}_i;\boldsymbol{\mu}_k,\boldsymbol{\Sigma}_k)}_{Q_2}, 
\end{equation}
with $y_{ik}^{(q)}= \esp_{\btheta^{(q)}} [Y_{ik} \mid \pmb{x}_i]$. 

We can observe that the  term $Q_2$ on right-hand side of \eqref{eq:Q} is identical to that of the auxiliary function for a standard GMM \eqref{eq:GMM1} with fixed proportions, for which parameter updates that maximizes $Q(\btheta,\btheta^{(q)})$ are known (and recalled below). The difference between our  EM procedure and the usual one for GMM's  thus resides in the maximization of the first term $Q_1$ with respect to the parameters defining the proportions $\pi_{ik}$. This procedure is detailed below.

\paragraph{E-step} 

In the E-step, we compute the conditional expectations $y_{tk}^{(q)}$ from the current parameter values as \cite{mclachlan1988mixture}:
\begin{equation}
y_{ik}^{(q)}= \frac{\phi(\pmb{x}_i;\boldsymbol{\mu}_k^{(q)},\boldsymbol{\Sigma}_k^{(q)}) \pi_{ik}^{(q)}}{\sum_{l=1}^K \phi(\pmb{x}_i;\boldsymbol{\mu}_l^{(q)},\boldsymbol{\Sigma}_l^{(q)}) \pi_{il}^{(q)}} \label{eq:ynkGMM}
\end{equation}
with $\phi$ given by \eqref{eq:gausspdf}.

\paragraph{M-step for $\bmu_k$ and $\bSigma_k$}

In the M-step, parameters $\boldsymbol{\mu}_k$ and $\boldsymbol{\Sigma}_k$ are first updated by maximizing $Q_2$. The update equations are \cite{mclachlan1988mixture}:
\begin{equation}
\boldsymbol{\mu}_k^{(q+1)}= \frac{1}{N_k} \sum_i y_{ik}^{(q)} \pmb{x}_i,
\end{equation}
with $N_k =  \sum_i y_{ik}^{(q)}$, and 
\begin{equation}
\boldsymbol{\Sigma}_k^{(q+1)}= \frac{1}{N_k} \sum_i y_{ik}^{(q)} ( \pmb{x}_i - \boldsymbol{\mu}_k^{(q+1)} )( \pmb{x}_i - \boldsymbol{\mu}_k^{(q+1)} )^\intercal.
\end{equation}


\paragraph{M-step for $\tau_k$, $\beta_k$ and $\gamma_k$}

Since no explicit update equations for parameters $\tau_k$, $\beta_k$ and $\gamma_k$ can be obtained, they need to be updated by an iterative optimization procedure. To enforce the constraints \eqref{eq:constr}, we first introduce the following auxiliary variables:
\begin{equation}
\tau_k=  \frac{T}{1+\exp(-\xi_k)}, \quad 
\beta_k = b_k^2,
\quad
\gamma_k =g_k^2.
\end{equation}
The calculation of the derivatives of $Q_1$ with respect to $\xi_k$, $b_k$ and $g_k$ is detailed in \ref{mzjkdpzakodoakda}. 
Using the gradient, we can then use any unconstrained nonlinear optimization procedure. In the experiments reported in Section \ref{sec:3}, we used a trust region algorithm implemented in the MATLAB 2020b Optimization toolbox. {{Several schemes were implemented and tested, included a Generalized EM \cite{ChatterjeeGEM} and  various optimization algorithms. The trust region method globally provided the best results.}} 


\subsubsection{Regularisation of the $\tau_k$'s}
\label{kkjddldzeefdzedz}

In the considered application related to acoustic emission data clustering, physical knowledge can be available suggesting when, in the timeline of a test, some damages must have occurred. This information can be provided, for example, as prior values $\tau_k^{\textrm{prior}}$ for $\tau_k$ (for some or all $k$ depending on the application). In this case, the auxiliary function \eqref{eq:Q} can be replaced by a  regularised version
\begin{equation}
\label{eq:Qbis}
    Q_r(\btheta,\btheta^{(q)}) = Q(\btheta,\btheta^{(q)}) - \lambda \Vert \boldsymbol{\tau} - \boldsymbol{\tau}^{\textrm{prior}} \Vert_2^2,
\end{equation}
where $\lambda$ is a regularization coefficient, $\|\cdot\|$ is the $L_2$ norm and $\boldsymbol{\tau}$ is the vector of $\tau_k$'s for which a prior value is available. {{In that case, the updating is modified for $\boldsymbol{\tau}$ according to Equation \ref{eq:Qr}.}}

\section{Experiments}
\label{sec:3}

The \textsf{GMMSEQ} method is first illustrated on a toy {{data set}} in Section \ref{subsec:toy}. It is then applied to real experimental data from a mechanical system in Section \ref{subsec:real}. 

\subsection{Simulated {{data set}}}
\label{subsec:toy}

\paragraph{Model for data generation}

A simulated {{data set}} was generated from the following model with $K = 4$ clusters:
\[
\boldsymbol{\mu}_1 = \begin{bmatrix} 1 & 1 \end{bmatrix}, \quad
\boldsymbol{\mu}_2 = \begin{bmatrix} 2 & 3 \end{bmatrix} , \quad
\boldsymbol{\mu}_3 = \begin{bmatrix} 3 & 5 \end{bmatrix} , \quad
\boldsymbol{\mu}_4 = \begin{bmatrix} 5 & 6 \end{bmatrix} ,
\]
and
\[
\boldsymbol{\Sigma}_1 = \begin{bmatrix} 0.3 & 0.2 \\ 0.2 & 0.2 \end{bmatrix} , \quad
\boldsymbol{\Sigma}_2 = \begin{bmatrix} 0.3 & 0.2 \\ 0.2 & 0.2 \end{bmatrix}, \quad
\boldsymbol{\Sigma}_3 = \begin{bmatrix} 0.2 & 0.1 \\ 0.1 & 0.3 \end{bmatrix}, \quad
\boldsymbol{\Sigma}_4 = \begin{bmatrix} 0.2 & 0.1 \\ 0.1 & 0.2 \end{bmatrix}.
\]
The numbers of observations in  the four clusters were set to $[1000, 1000, 3000, 1000]$
and the timestamps were generated  randomly  from a uniform distribution
$$t_i = t_{i-1} + \mathcal{U}_{[0,1]},$$ 
starting from $t_1 = 0$.
The value of $T$ is thus $\max_i t_i$ and the length of the data is equal to $6000$.
The parameters defining the time-varying proportions were set as follows:
\[
\boldsymbol{\beta} = \begin{bmatrix} 2.72 & 10.1 & 30.2  \end{bmatrix}, \quad
\boldsymbol{\gamma} = \begin{bmatrix} 0.009 & 0.015 & 0.012 \end{bmatrix}, 
\]    
and the $\tau_k$ were initialized to the time stamps $t_i$ with $i\in\{488, 1990, 2472\}$. 





\begin{figure}
     \centering
     \begin{subfigure}[b]{0.49\textwidth}
         \centering
         \includegraphics[width=1\textwidth]{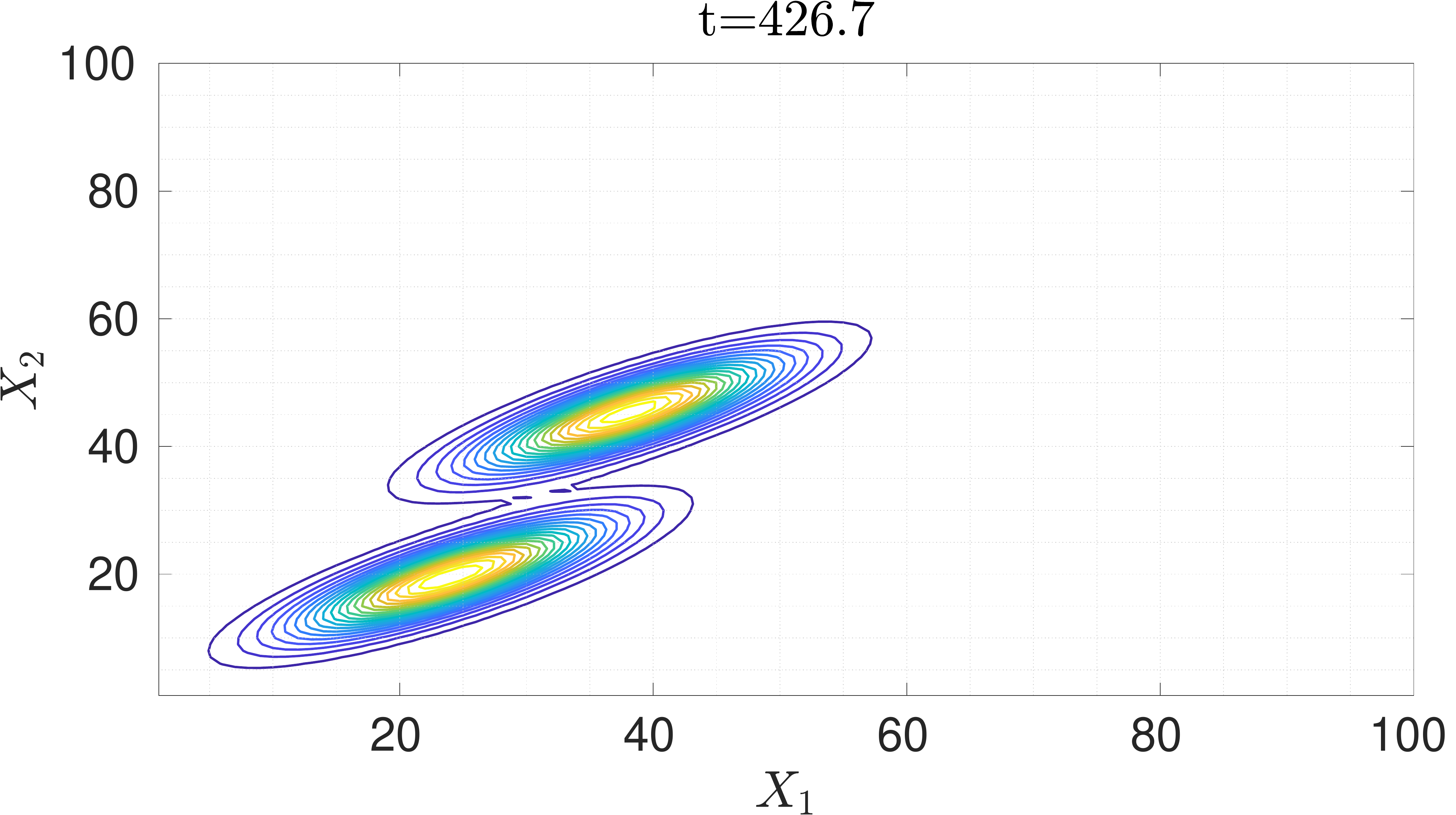}
         \caption{}
         \label{fig:f1seq}
     \end{subfigure}
     \hfill
     \begin{subfigure}[b]{0.49\textwidth}
         \centering
         \includegraphics[width=1\textwidth]{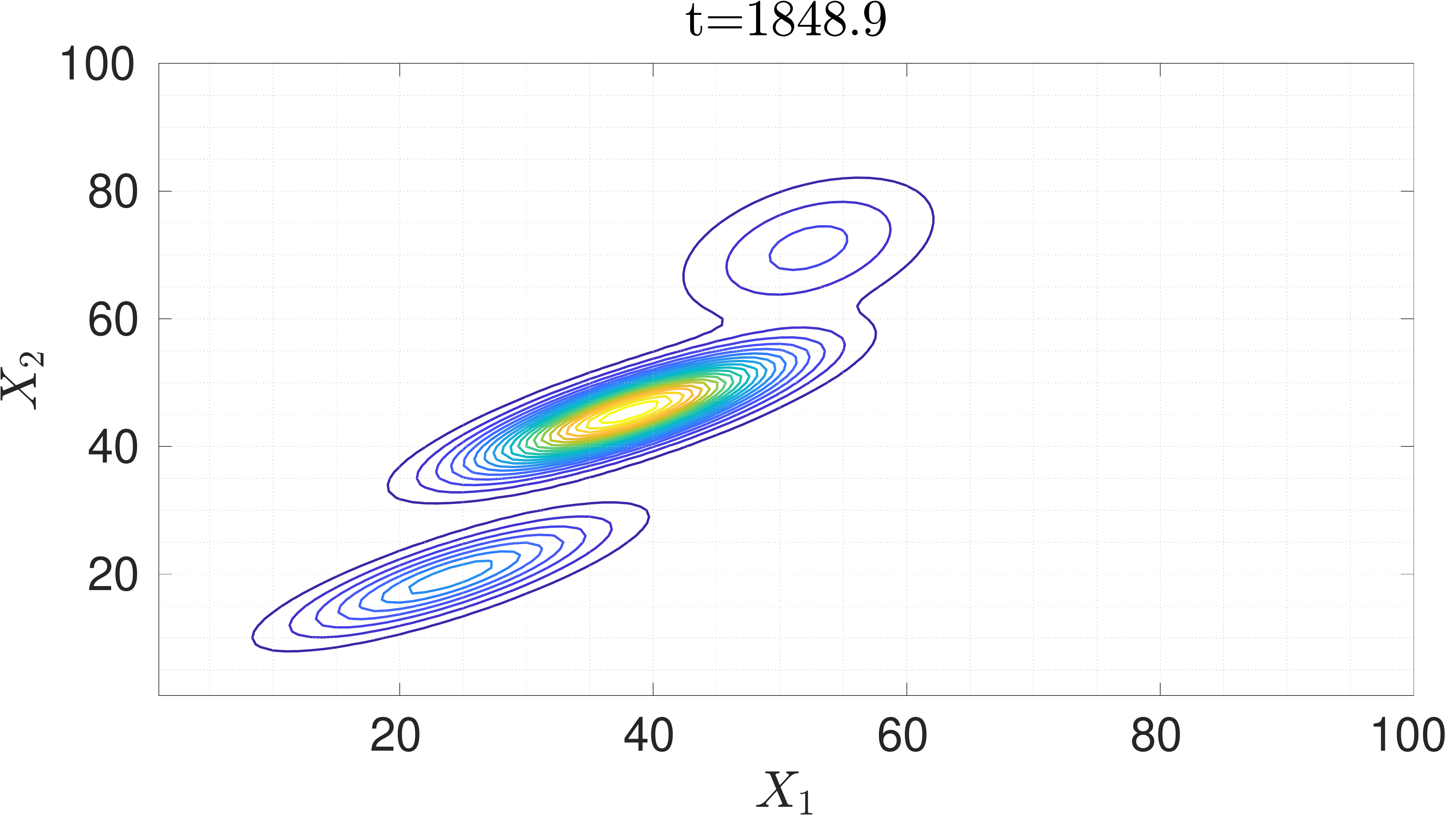}
         \caption{}
         \label{fig:f2seq}
     \end{subfigure}
     \hfill
     \begin{subfigure}[b]{0.49\textwidth}
         \centering
         \includegraphics[width=1\textwidth]{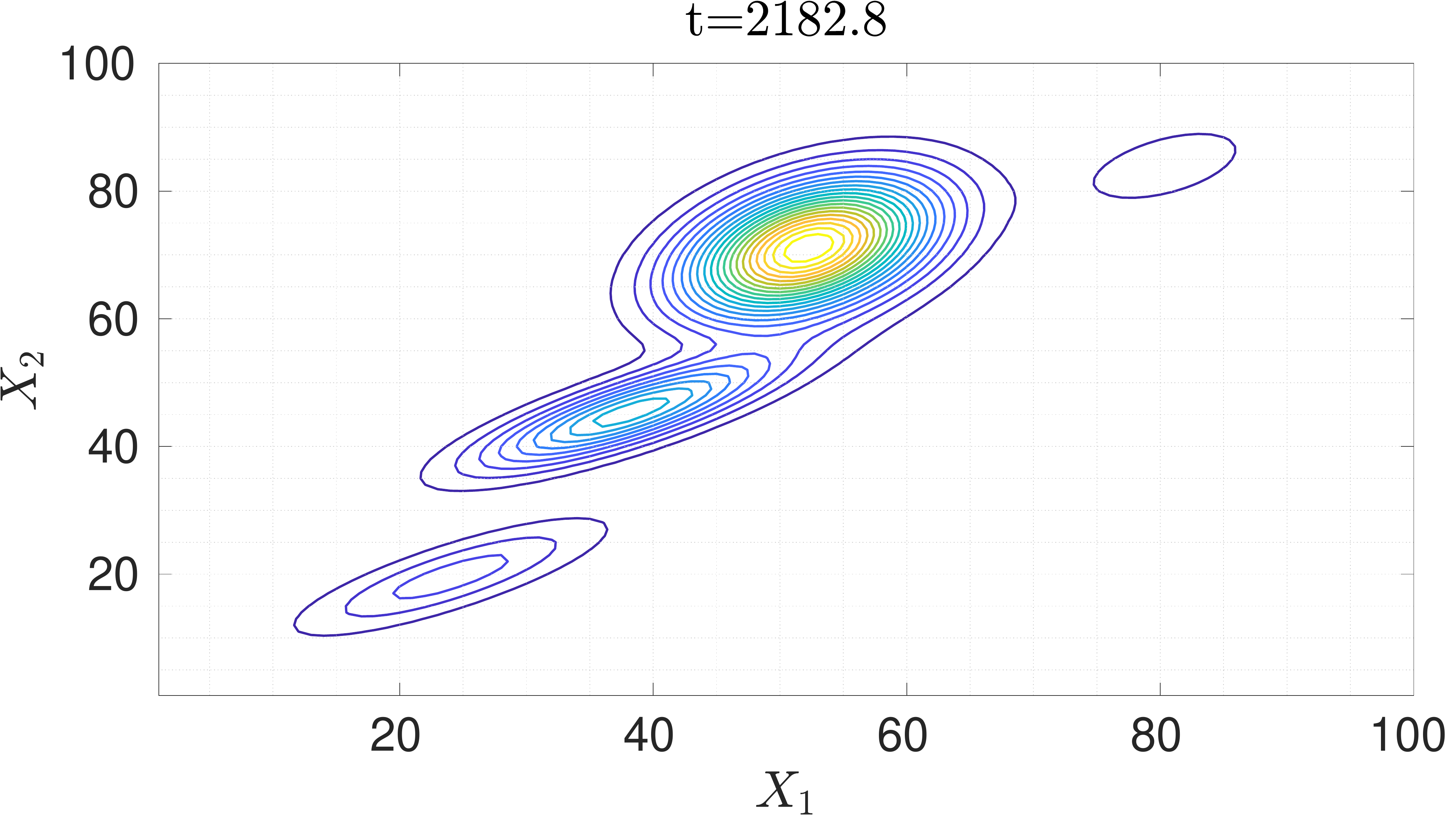}
         \caption{}
         \label{fig:f3seq}
     \end{subfigure}
     \begin{subfigure}[b]{0.49\textwidth}
         \centering
         \includegraphics[width=1\textwidth]{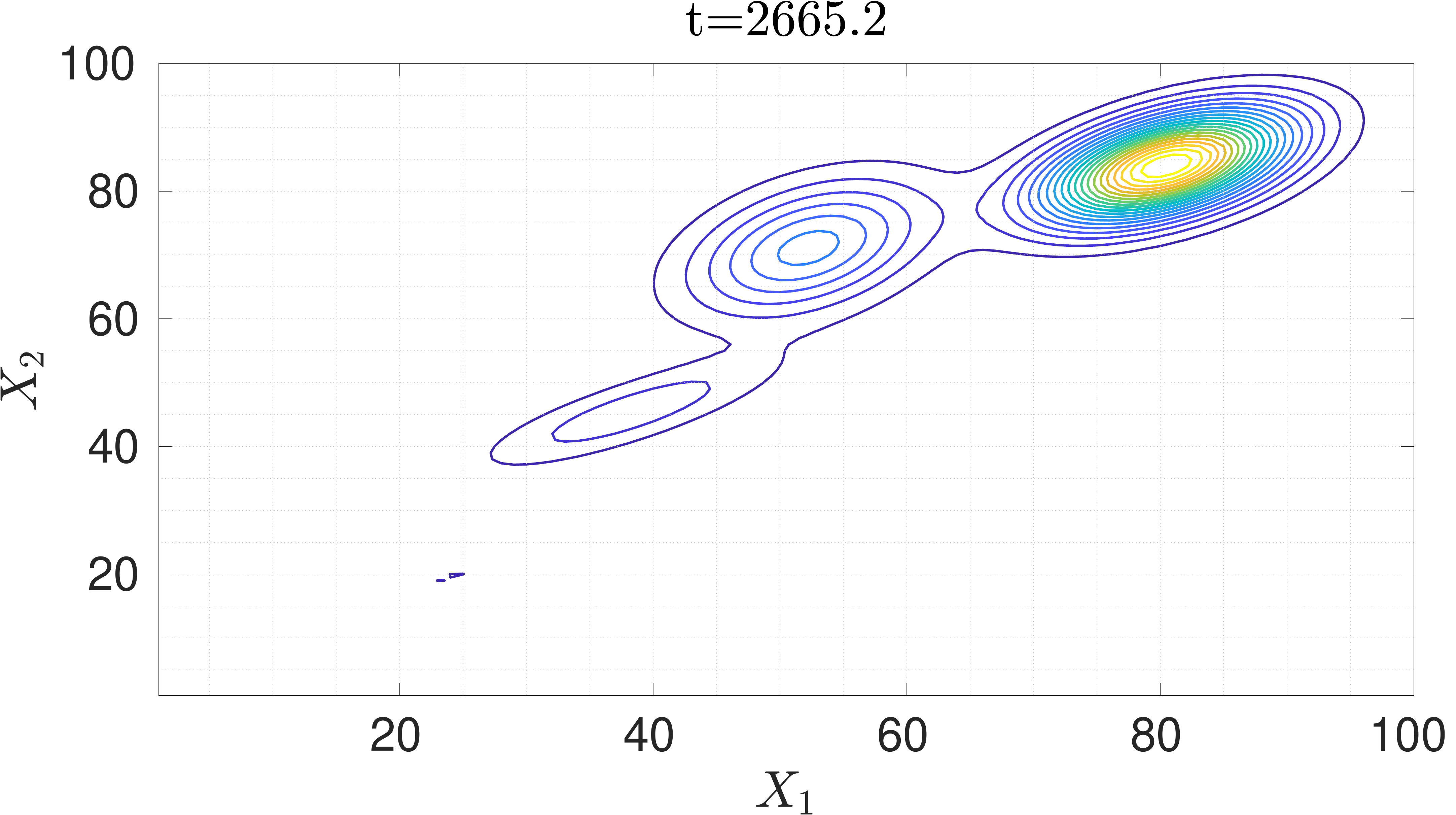}
         \caption{}
         \label{fig:f4seq}
     \end{subfigure}
        \caption{Contours of the mixture density at four successive time steps.  
        The corresponding proportions $\pi_{ik}$ correspond to the locations of the square  markers in Figure~\ref{fig:ffezlkjfekzlejfzfzfzf}.
        \label{fig:dataSimuGMMseq}}
\end{figure}

\begin{figure}
\centering
\includegraphics[width=0.95\textwidth]{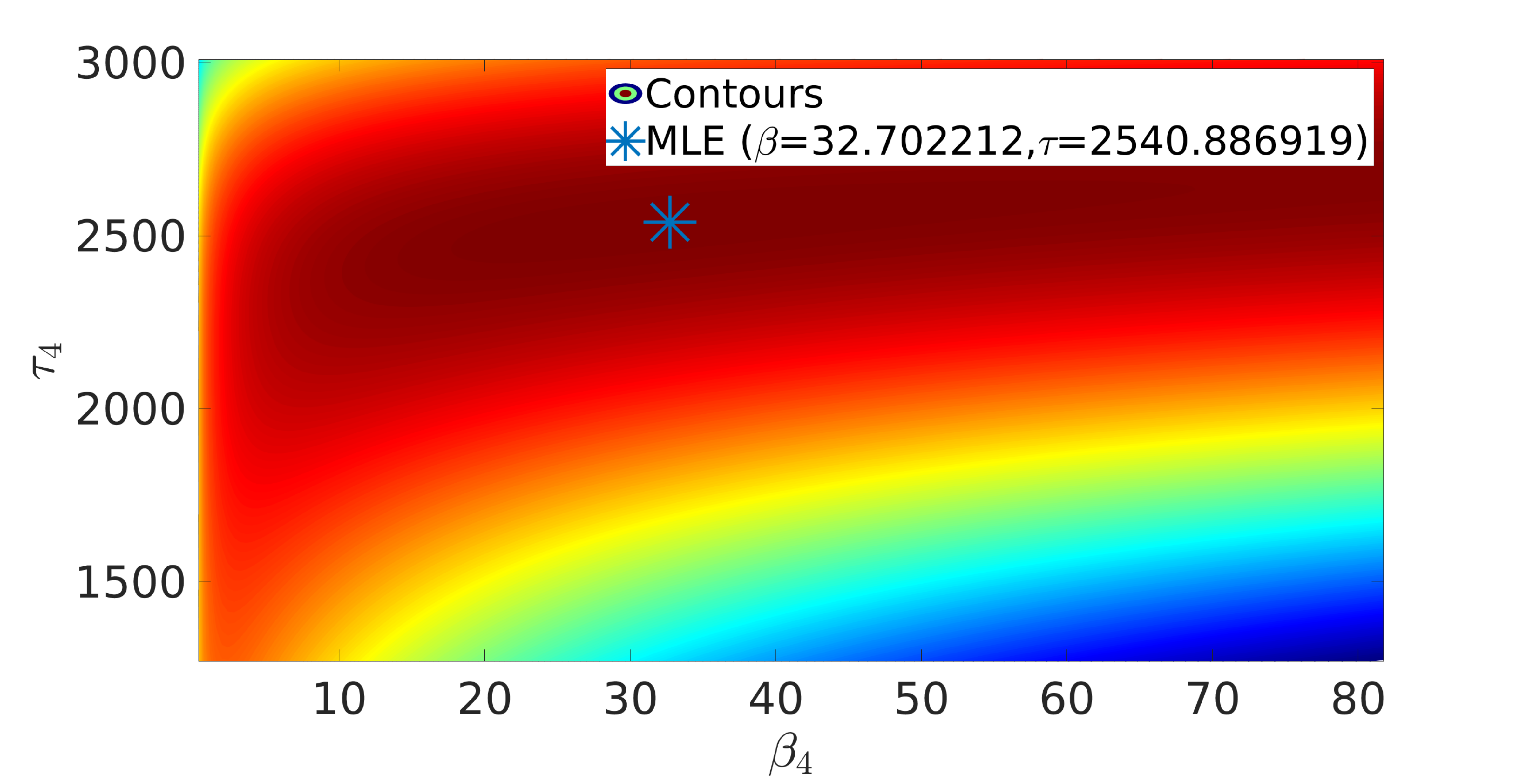}
\caption{Contours of the log-likelihood in the plane $(\beta_4,\gamma_4)$, the other parameters being fixed to their maximum likelihood estimates. \label{zdzzdzefzfz22}}
\end{figure}

Because the proportions in this model vary with time, the mixture density also depends on time. Contours of the mixture density are depicted in Figure \ref{fig:dataSimuGMMseq} at four different time steps  showing the gradual emergence of the four clusters. Figure \ref{zdzzdzefzfz22} shows a contour plot of the likelihood function (assuming the correct number of clusters) as a function of $\beta_4$ and $\gamma_4$, with the other parameters fixed at their maximum likelihood estimates.

\paragraph{Model selection}

\begin{figure}
             \centering
         \includegraphics[width=0.8\textwidth]{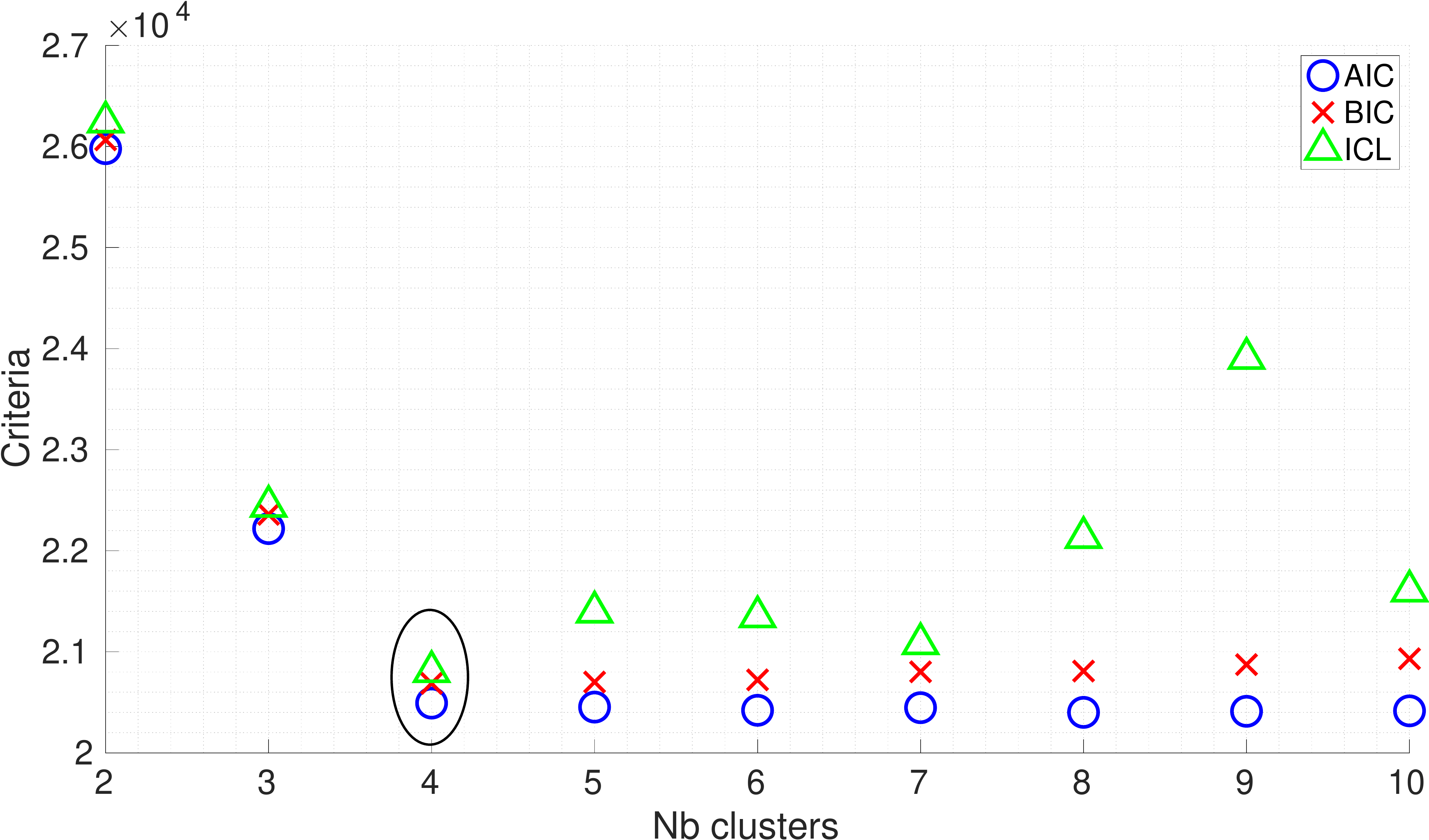}
     \caption{AIC, BIC and ICL of \textsf{GMMSEQ} models with different numbers of clusters for the simulated data.}
        \label{fig:ldjoijdoijfdijd}
\end{figure}

In practice, we need a criterion allowing us to select the number of clusters automatically. For mixture models,  common choices are the Akaike Information Criterion (AIC), the Bayesian Information Criterion (BIC) and the Integrated Completed Likelihood (ICL) \cite{biernacki1999choosing,baudry2010combining}. We used these three criteria  to select the best \textsf{GMMSEQ} model for the simulated data. For each run, the model was initialized by: 
\begin{enumerate}
\item A standard GMM (with a Matlab 2020b implementation provided in the Statistics and Machine Learning Toolbox);
\item The K-means  algorithm (provided in the same toolbox as for GMM);
\item A segmentation of the data into $K$ blocks of the same size and a Gaussian distribution fitted to each block.
\end{enumerate}
For the three types of initialization, 10 runs were performed for a number of clusters varying from $2$ to $10$. After convergence, for each number of clusters, the best run was selected according to the value of likelihood computed by \eqref{eq:GMM2}. The values of  AIC, BIC and ICL were finally computed for each model. The criteria are plotted against the number $K$ of clusters in Figure \ref{fig:ldjoijdoijfdijd}. AIC shows an evolution presenting an ``elbow'' from which the number of clusters can be chosen, while BIC and ICL have a minimum for the correct number of clusters ($K=4$).  For ICL, depending on the runs, the evolution can show several local minima as in the figure or a single one, but the global minimum is always located at the correct value. For different runs,  BIC and AIC showed a consistent elbow-shaped behavior with a minimum that can be more or less pronounced for the correct number of clusters. This study with simulated data suggests that the three criteria have the ability to provide the correct number of clusters. 

\paragraph{Onset estimation}

\begin{figure}
     \centering
     \begin{subfigure}[b]{0.7\textwidth}
         \centering
         \includegraphics[width=1\textwidth]{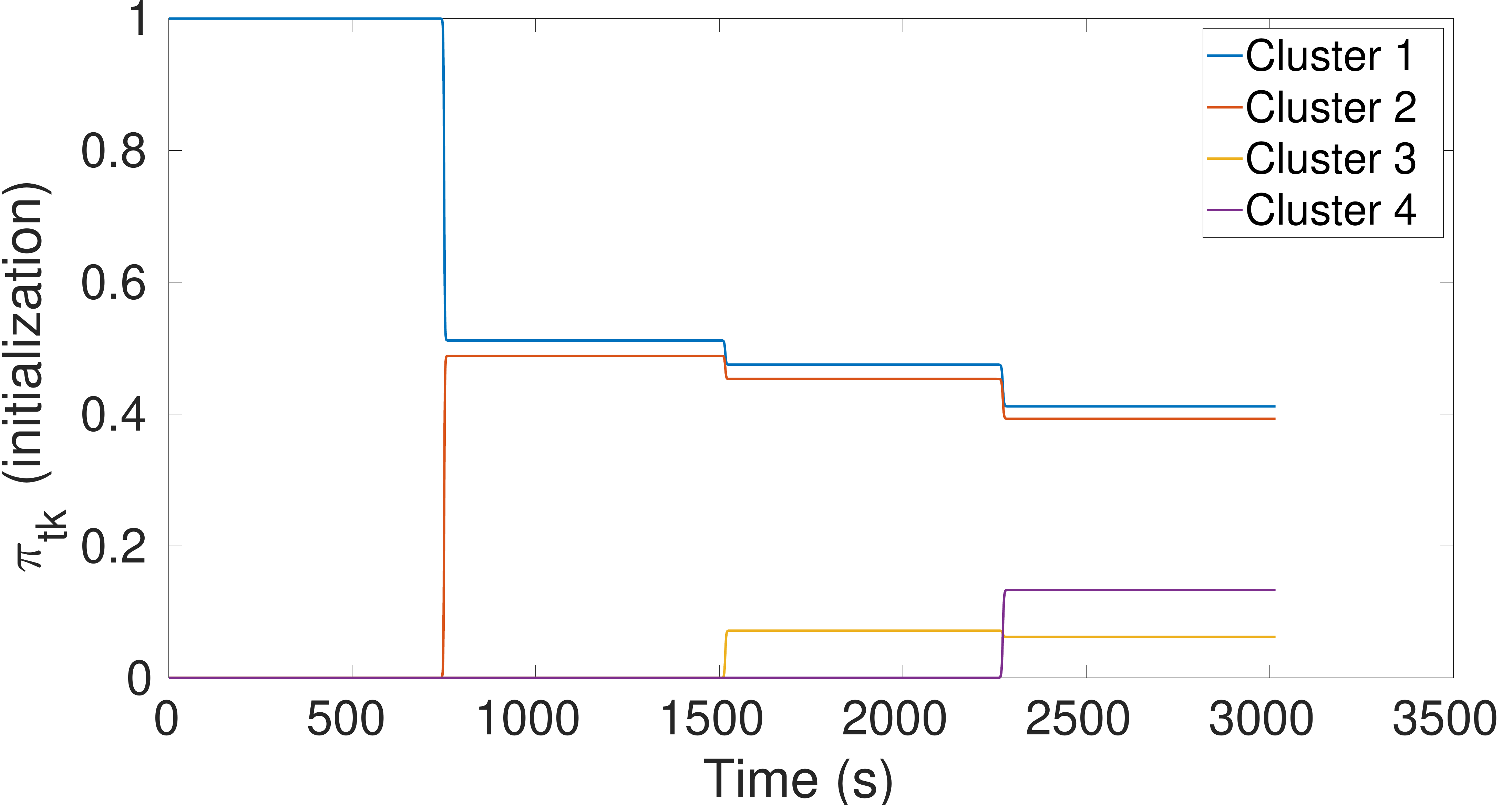}
         \caption{Initialization.}
     \end{subfigure}
     \begin{subfigure}[b]{0.7\textwidth}
         \centering
         \includegraphics[width=1\textwidth]{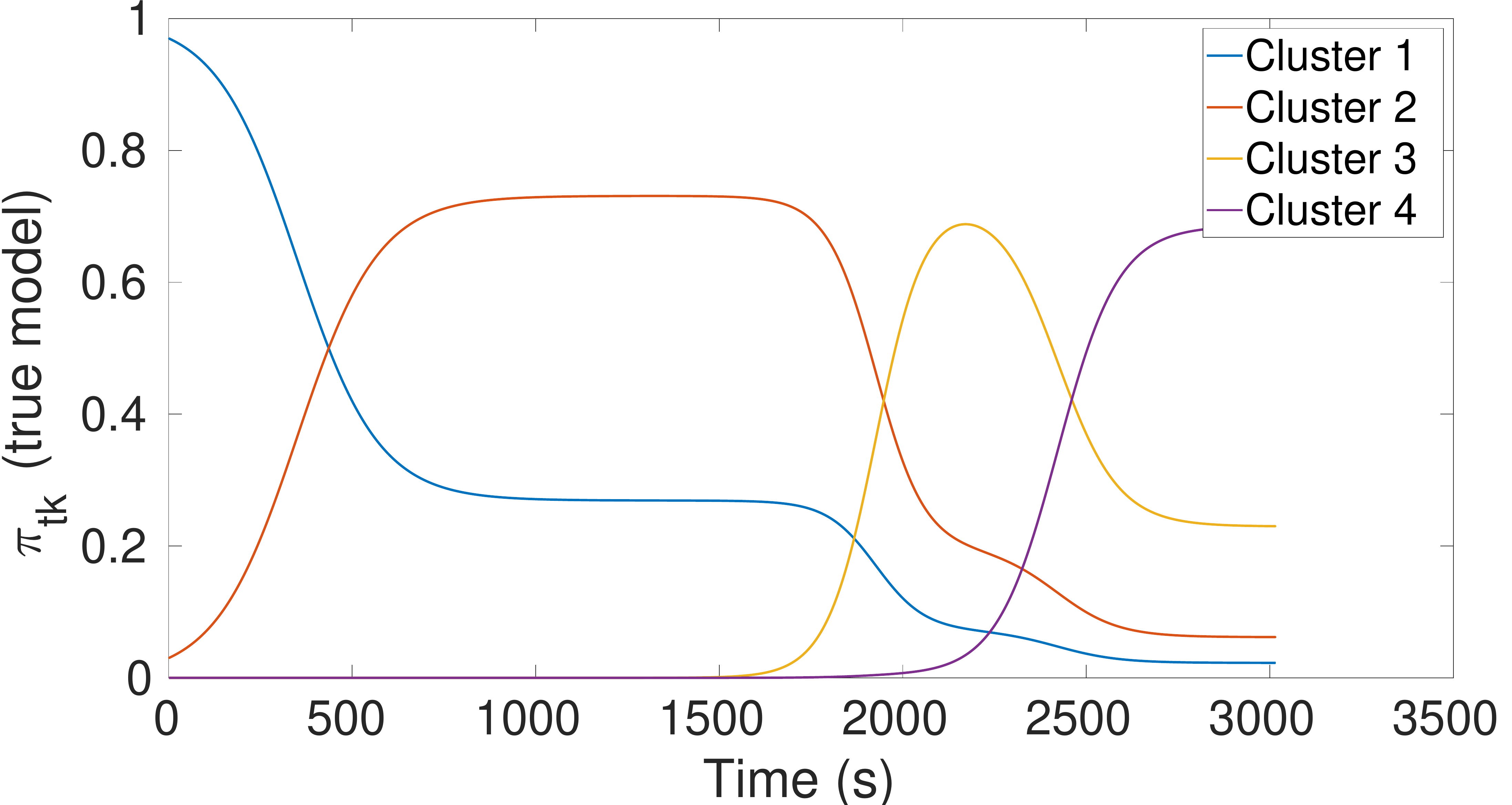}
         \caption{True values.}
     \end{subfigure}
     \begin{subfigure}[b]{0.7\textwidth}
         \centering
         \includegraphics[width=1\textwidth]{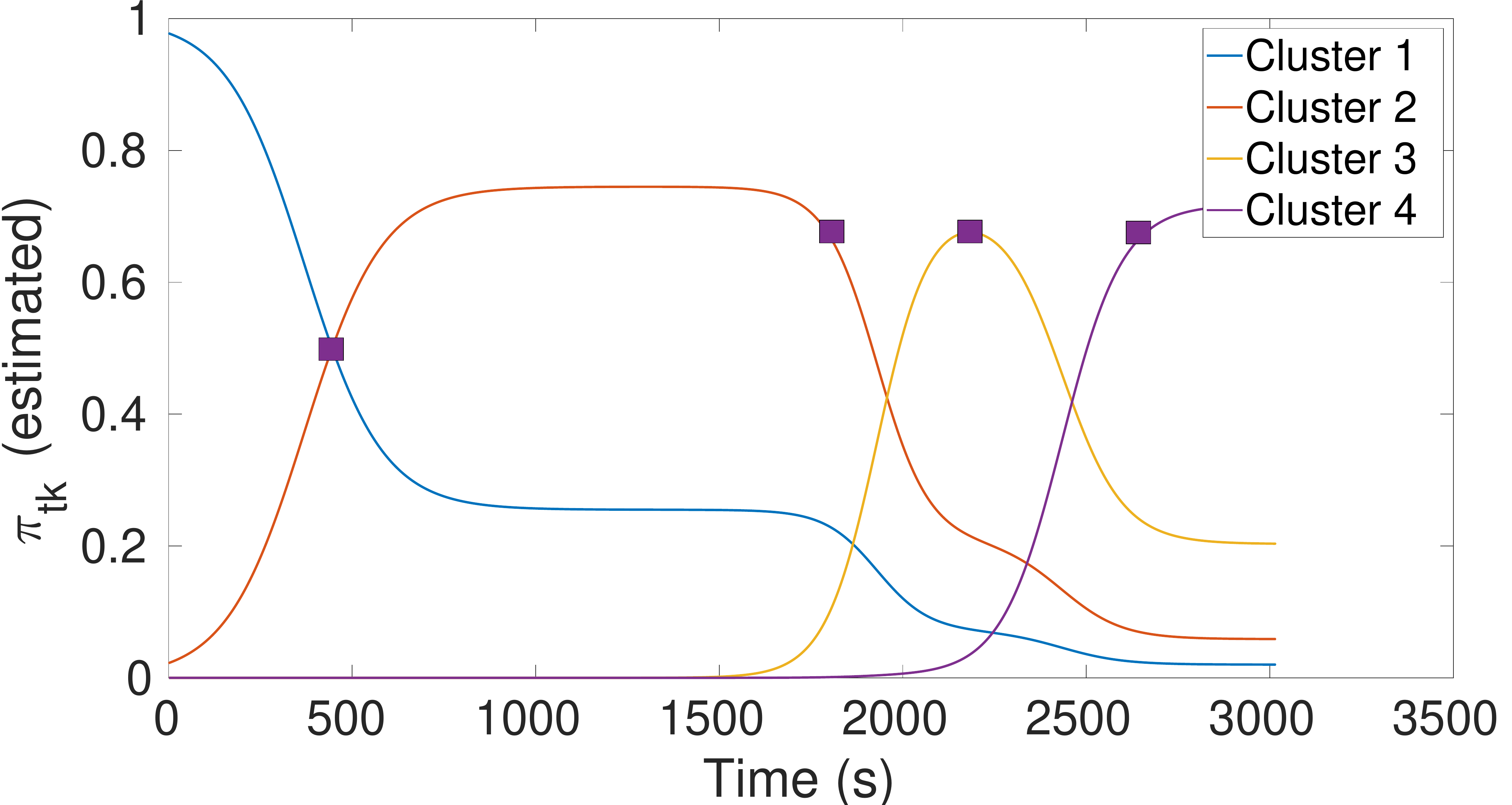}
         \caption{Estimated values.}
     \end{subfigure}    
        \caption{Evolution of the proportions $\pi_{ik}$ as a function of time (in arbitrary units and was generated randomly as explained in the text). Squared-shape markers represent the values used in Figure~\ref{fig:dataSimuGMMseq}.}
        \label{fig:ffezlkjfekzlejfzfzfzf}
\end{figure}

In the introduction, we motivated this work by the need to process data with continuous timestamps and we proposed a model including an estimation of the onsets of clusters. The time-varying proportions ($\pi_{ik}$) are represented in Figure \ref{fig:ffezlkjfekzlejfzfzfzf} using the true values of parameters, the estimated ones, and the values used in the initialization of \textsf{GMMSEQ}. We can observe that {{the estimated proportions ($\boldsymbol{\beta}$),  kinetics parameter ($\boldsymbol{\gamma}$) and onsets ($\boldsymbol{\tau}$) are very close to the true ones, validating the identification procedure in \textsf{GMMSEQ}.}} 

\subsection{Real data}
\label{subsec:real}

\paragraph{{{Data set}} description}

\begin{figure}
\centering
\includegraphics[width=0.9\linewidth]{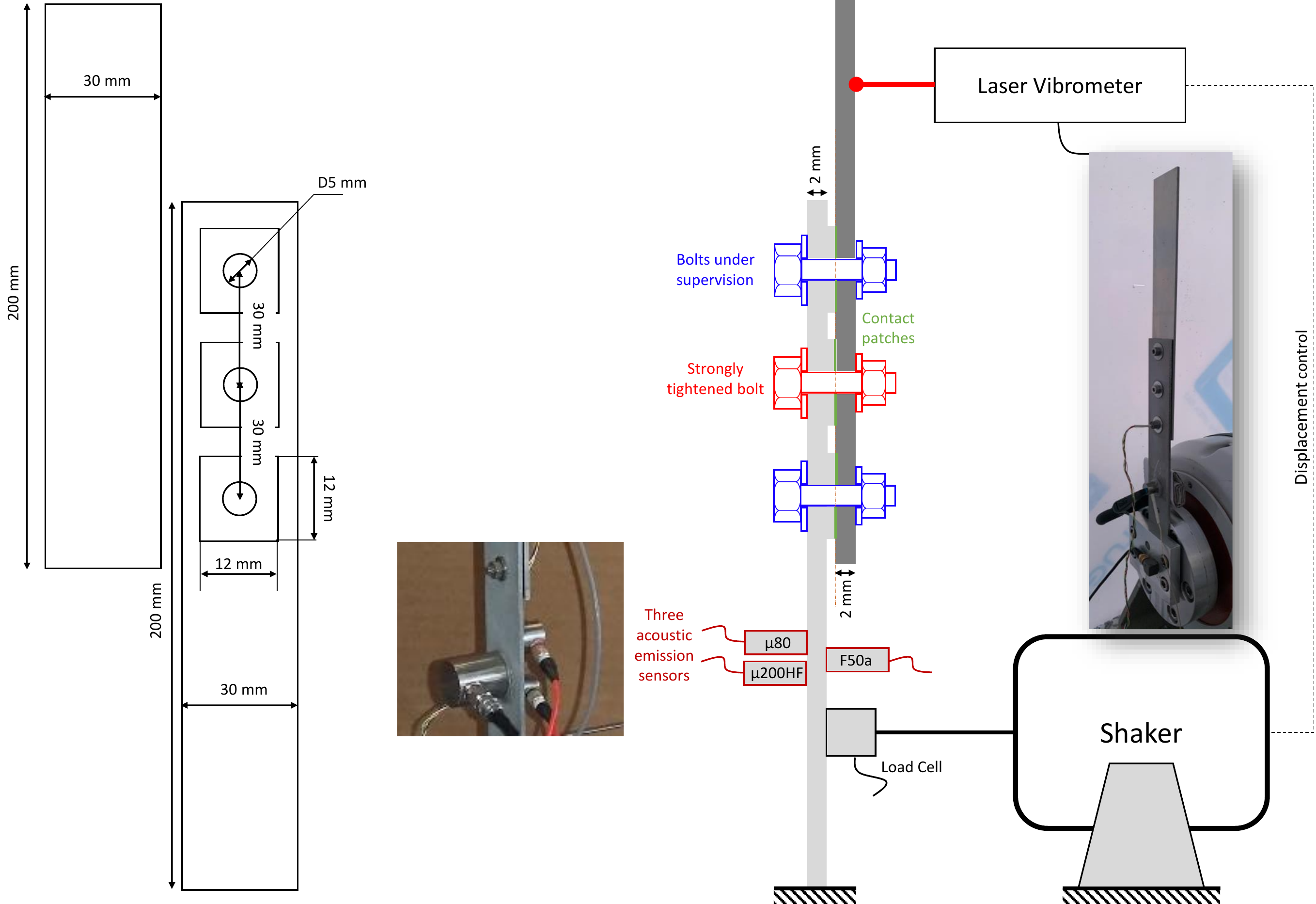}
\caption{Setup description: part dimensions, sensors and bolts position (from \cite{ORIONdata}). \label{fig:setup}}
\end{figure}

The benchmark {{data set}} ORION-AE \cite{orionaedata,ORIONdata} is used in this section to demonstrate the performance of the \textsf{GMMSEQ} method. The experiments were designed  to reproduce the loosening phenomenon observed in aeronautics, automotive or civil engineering structures where parts are assembled together by means of bolted joints (Figure \ref{fig:setup}). The bolts can, indeed, be subject to self-loosening under vibrations. Consequently, it is of paramount importance to develop sensing strategies and algorithms for early loosening estimation \cite{zhang2019continuous}.

The ORION-AE {{data set}} is composed of five parts collected during five measurement campaigns  denoted as $B$, $C$, $D$, $ E$ and $F$ in the sequel. ORION is a simple jointed structure made of two plates manufactured in a 2024 aluminium alloy, linked together by three bolts. The contact between the plates is done through machined overlays. The contact patches have an area of $12\times 12$ mm$^2$ and are {{$2$ mm}} thick. The structure was submitted to a $100$ Hz harmonic excitation force. The load was applied using a Tyra electromagnetic shaker, which can deliver a $200$ N force. The force was measured using a PCB piezoelectric load cell and the vibration level was determined next to the end of the specimen using a Polytec laser vibrometer. 

Seven  tightening levels were applied on the upper bolt. The tightening was first set  to $60$ cNm with a torque screwdriver. After a $10$ seconds vibration test, the shaker was stopped and this vibration test was repeated after a torque modification at $50$ cNm. Torque modifications at $40$, $30$, $20$, $10$ and $5$ cNm were then applied. Note that, for campaign $C$, the level $20$ cNm is missing. All dimensions are detailed in  Figure \ref{fig:setup} to enable readers to reproduce the test. 

For each campaign, four sensors were used: a laser vibrometer and three different AE sensors (micro-200-HF, micro-80 and the F50A from Euro-Physical Acoustics) with various frequency bands were attached onto the lower plate. All data were sampled at 5 MHz. The velocimeter was used to control the amplitude of the displacement of the top of the upper beam so that it remains constant for all tightening levels. During vibrations, stick-slip transitions or shocks in the interface generate small AE events which are dependent on bolt tightening. These sources of AE signals have to be detected and identified from the data stream, which constitutes the challenge. 

\begin{figure}[!ht]
\centering
\includegraphics[width=0.9\linewidth]{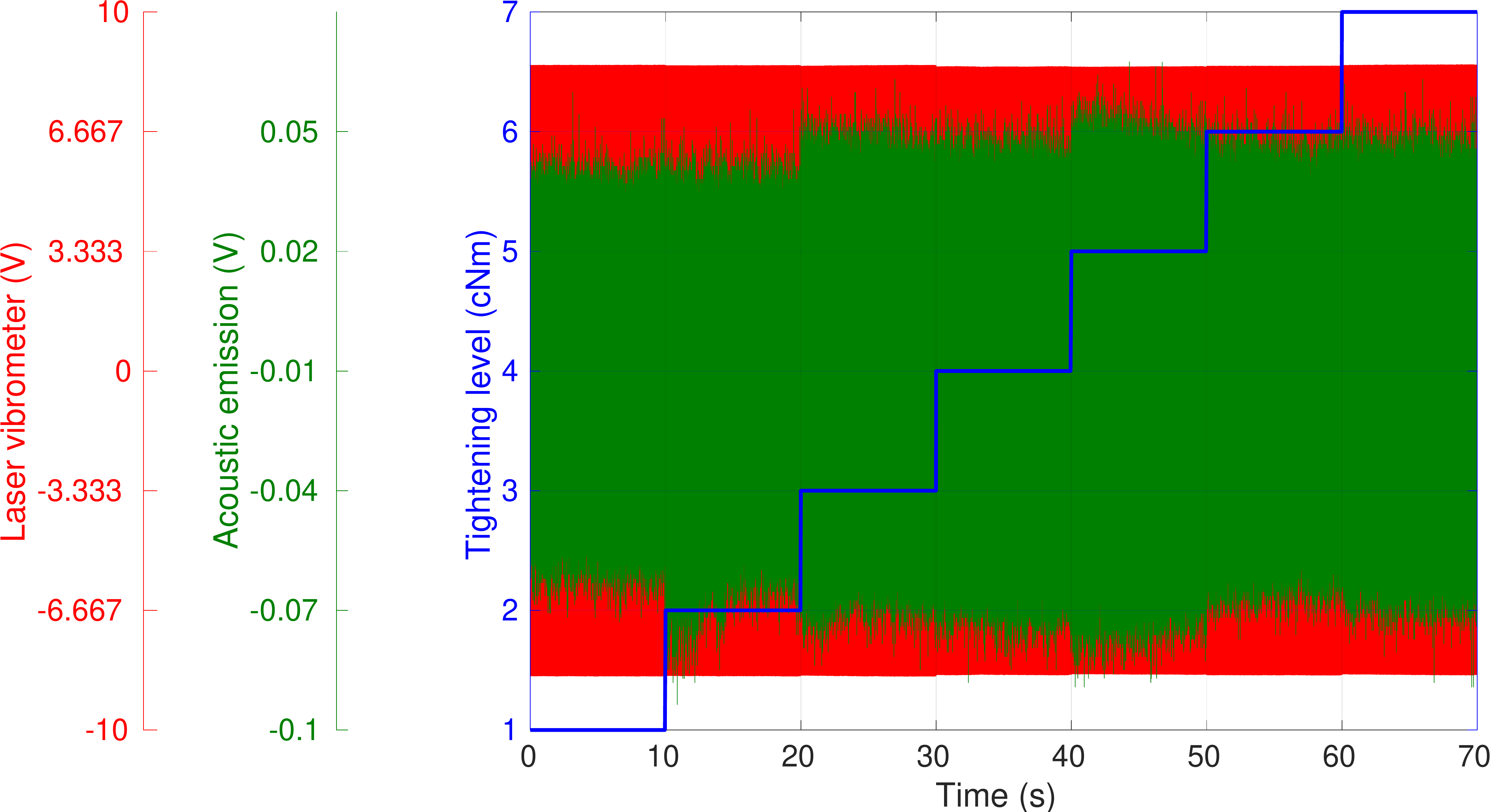}
\caption{{{Tightening levels, acoustic emission and laser vibrometer data superimposed for measurements ``F'' and sensor micro-200-HF. The x-axis is here represented using the time of test (about $70$ s), starting from $60$ cNm from the left (around between $t \in [0,10]$ s) to $5$ cNm on the right (from $t > 60$ s). }}
\label{fzefdfffzefzef33133}}
\end{figure}

AE {{data sets}} are generally unlabeled because it is not possible to identify the AE source with certainty {{for all AE signals.}} {{However, the ORION-AE data set contains raw data for which the tightening levels are known. Therefore, it represents a good case study for performance benchmarking of clustering methods like \textsf{GMMSEQ} or existing methods like GMM, K-means, hierarchical clustering (HC) and Gustafson-Kessel (GK) algorithms}.}
{{The data set is presented in a companion paper \cite{ORIONdata} with illustrations of raw data and signal processing for different campaigns. Figure \ref{fzefdfffzefzef33133}  depicts the data in campaign F where the green curve represents the raw AE data (as used in this study). The stairstep curve in blue represents the tightening levels. The red curve corresponds to the vibrometer data (reflecting harmonic vibration at 100 Hz with displacement control). }} 

\paragraph{Signal processing}

\begin{figure}
\centering
\includegraphics[width=0.5\textwidth]{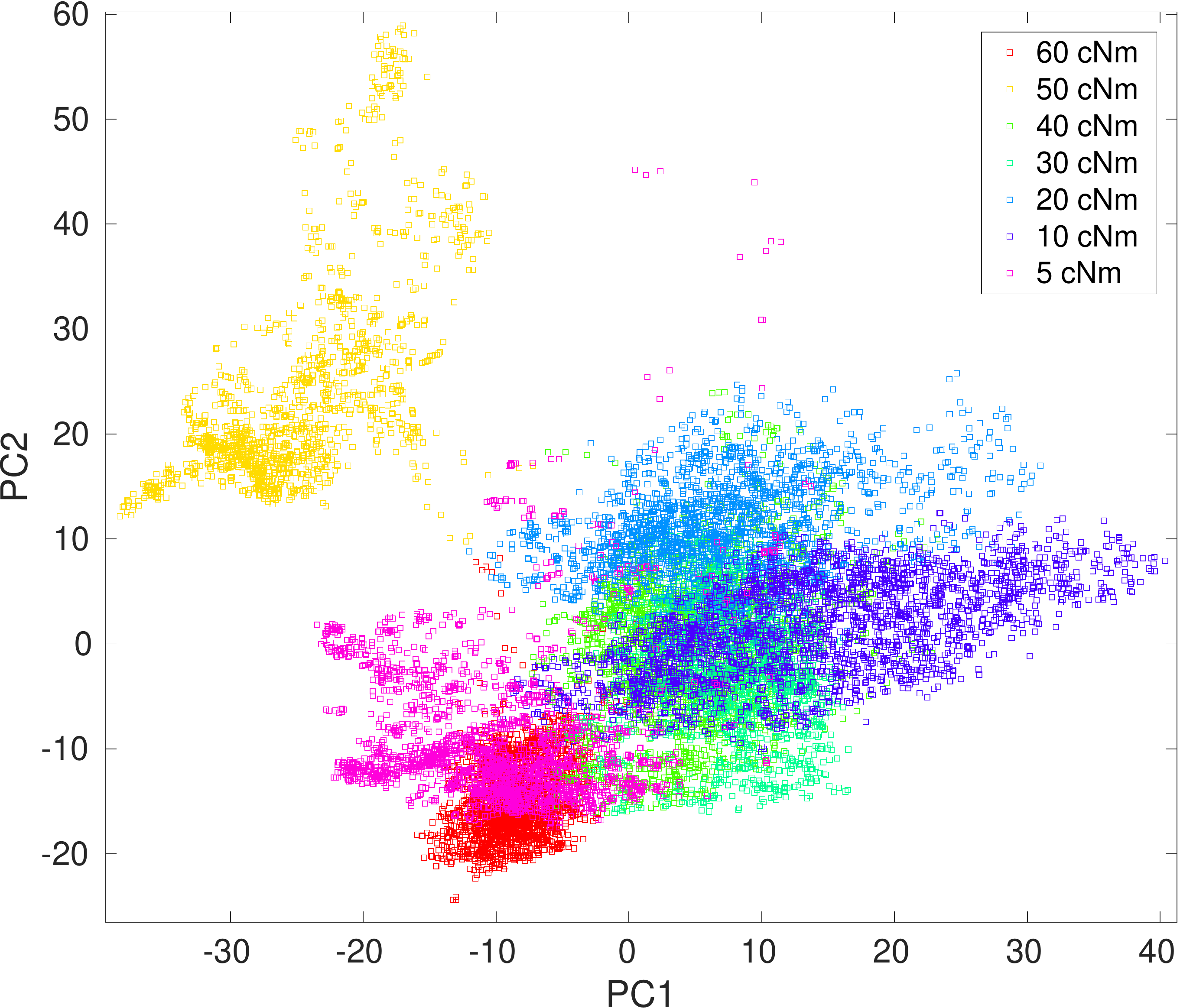}
\caption{First two principal components of campaign $E$ data. \label{ldkpozikedopz}}
\end{figure}

{{When using a feature-based clustering algorithm like \textsf{GMMSEQ}, the raw AE data must be first preprocessed by a hit detection procedure. This procedure is common to most  AE data analyses and therefore, it is not often described in publications because commercial softwares are often used. In this work, the method used was described in \cite{kharrat2016signal}. It is summarized in \ref{kharratetal} and some illustrations of this method applied to the ORION-AE data can also be found in \cite{ORIONdata}. The method provides the following features, which are commonly used in AE literature \cite{pcimistras,sause12,kharrat2016signal,Kattis17}}}: Rise time, counts, PAC-energy, duration, amplitude, average frequency, RMS, average signal level, counts to peak, reverberation frequency, initiation frequency, signal strength, absolute energy, partial power in the intervals $[0, 20, 100, 200, 300, 400, 500, 600, 800, 1000]$ kHz, frequency centroid, peak frequency, weighted peak frequency. To this set of features were added the following ones: the Renyi number calculated from the scalogram as in \cite{gonzalez2000measuring} using a Morlet wavelet, as well as the frequency of the maximum of energy in the scalogram\footnote{{{The feature matrices for all campaigns are available at \url{https://drive.google.com/drive/folders/1H413RxYu4ya7YMEgF_lTh_fHr7flvvOO?usp=sharing}.}}}. The set of feature vectors obtained in each campaign were then postfiltered using a 31-sample moving median applied to each dimension of the resulting feature matrix in order to ensure temporal coherence. Principal Components Analysis (PCA) was then used to extract the $n$ first components explaining $99 \%$ of the variance. The value of $n$ varies for the different campaigns as shown in Table \ref{mlksapzkosa}, where sensor micro-200-HF was used. Figure \ref{ldkpozikedopz} displays the first two components for campaign $E$ and sensor micro-200-HF, where the colors are related to the level of loosening. 


\begin{table}[ht]
\caption{Some statistics about the feature extraction step (sensor $\mu 200HF$).\label{mlksapzkosa}}
\begin{center}
\begin{tabular}{lccccc}
\hline
Campaign & B & C & D & E & F \\
\hline
\# of tightening levels & $7$ & $6$ & $7$ & $7$ & $7$ \\
\# features before PCA & $32$ & $32$ & $32$ & $32$ & $32$ \\
\# features after PCA & $16$ & $17$ & $16$ & $23$ & $25$ \\
Total \#  of signals & $10,866$ & $9,461$ & $9,285$ & $15,628$ & $17,810$ \\
Average \#  of signals per period & $1.55$ & $1.57$ & $1.33$ & $2.23$ & $2.54$ \\
\hline
\end{tabular}
\end{center}
\end{table}
      
\subsection{Results}

We ran the algorithm 10 times for each of  the same three initialization methods mentioned in Section \ref{subsec:toy}. The  number of clusters was varied from $4$ to $14$. For each number of clusters, the parameter estimates corresponding to the highest likelihood were selected.

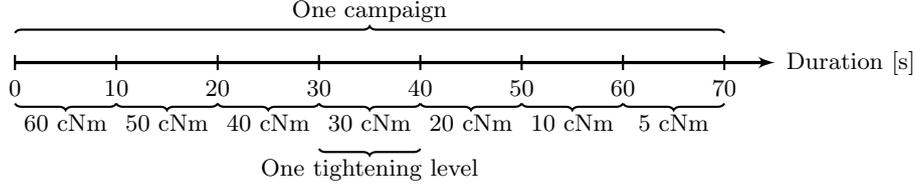
\begin{figure}[ht]
\centering
\begin{tikzpicture}[y=1cm, x=1cm, thick, font=\footnotesize]    
\usetikzlibrary{arrows,decorations.pathreplacing}

\tikzset{
   brace_top/.style={
     decoration={brace},
     decorate
   },
   brace_bottom/.style={
     decoration={brace, mirror},
     decorate
   }
}

\draw[line width=1.2pt, ->, >=latex'](0,0) -- coordinate (x axis) (10,0) node[right] {Duration [s]}; 
\foreach \x in {0,10,20,30,40,50,60,70} \draw (\x/10/0.75,0.1) -- (\x/10/0.75,-0.1) node[below] {\x};

\node (startC) at (0/0.75,0.25) {};
\node (endC) at (7/0.75,0.25) {};
\draw [brace_top] (startC.north) -- node [above, pos=0.5] {One campaign} (endC.north);

\node (startP) at (3/0.75,-1.0) {};
\node (endP) at (4/0.75,-1.0) {};
\draw [brace_bottom] (startP.south) -- node [below, pos=0.5] {One tightening level} (endP.south);

\node (startP) at (0/0.75,-0.4) {};
\node (endP) at (1/0.75,-0.4) {};
\draw [brace_bottom] (startP.south) -- node [below, pos=0.5] {60 cNm} (endP.south);

\node (startP) at (1/0.75,-0.4) {};
\node (endP) at (2/0.75,-0.4) {};
\draw [brace_bottom] (startP.south) -- node [below, pos=0.5] {50 cNm} (endP.south);

\node (startP) at (2/0.75,-0.4) {};
\node (endP) at (3/0.75,-0.4) {};
\draw [brace_bottom] (startP.south) -- node [below, pos=0.5] {40 cNm} (endP.south);

\node (startP) at (3/0.75,-0.4) {};
\node (endP) at (4/0.75,-0.4) {};
\draw [brace_bottom] (startP.south) -- node [below, pos=0.5] {30 cNm} (endP.south);

\node (startP) at (4/0.75,-0.4) {};
\node (endP) at (5/0.75,-0.4) {};
\draw [brace_bottom] (startP.south) -- node [below, pos=0.5] {20 cNm} (endP.south);

\node (startP) at (5/0.75,-0.4) {};
\node (endP) at (6/0.75,-0.4) {};
\draw [brace_bottom] (startP.south) -- node [below, pos=0.5] {10 cNm} (endP.south);

\node (startP) at (6/0.75,-0.4) {};
\node (endP) at (7/0.75,-0.4) {};
\draw [brace_bottom] (startP.south) -- node [below, pos=0.5] {5 cNm} (endP.south);

\end{tikzpicture}
\caption{Timeline of tightening levels to interpret the next figures. \label{ljklkjflzjelkf}}
\end{figure}

\begin{figure}
\centering
\includegraphics[width=0.75\textwidth]{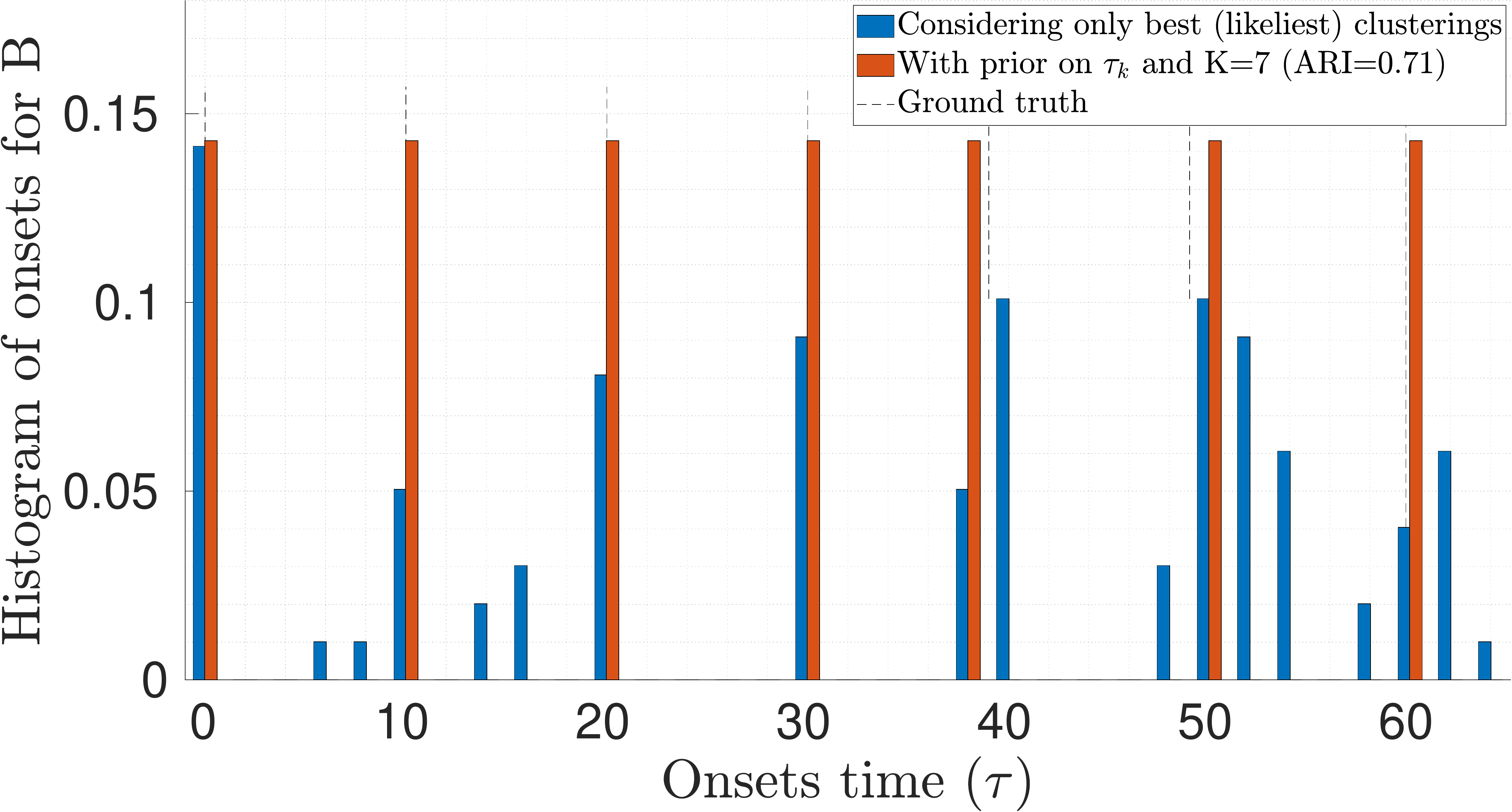}
\caption{Campaign B: (\textcolor{blue}{blue}) Histogram of $\tau_k$ estimates for the three initialization methods and $K$ ranging from 4 to 14, and (\textcolor{red}{red}) histogram with prior on onsets with regularization. \label{kqldkdkazlk}}
\end{figure}

\begin{figure}
\centering
\includegraphics[width=0.75\textwidth]{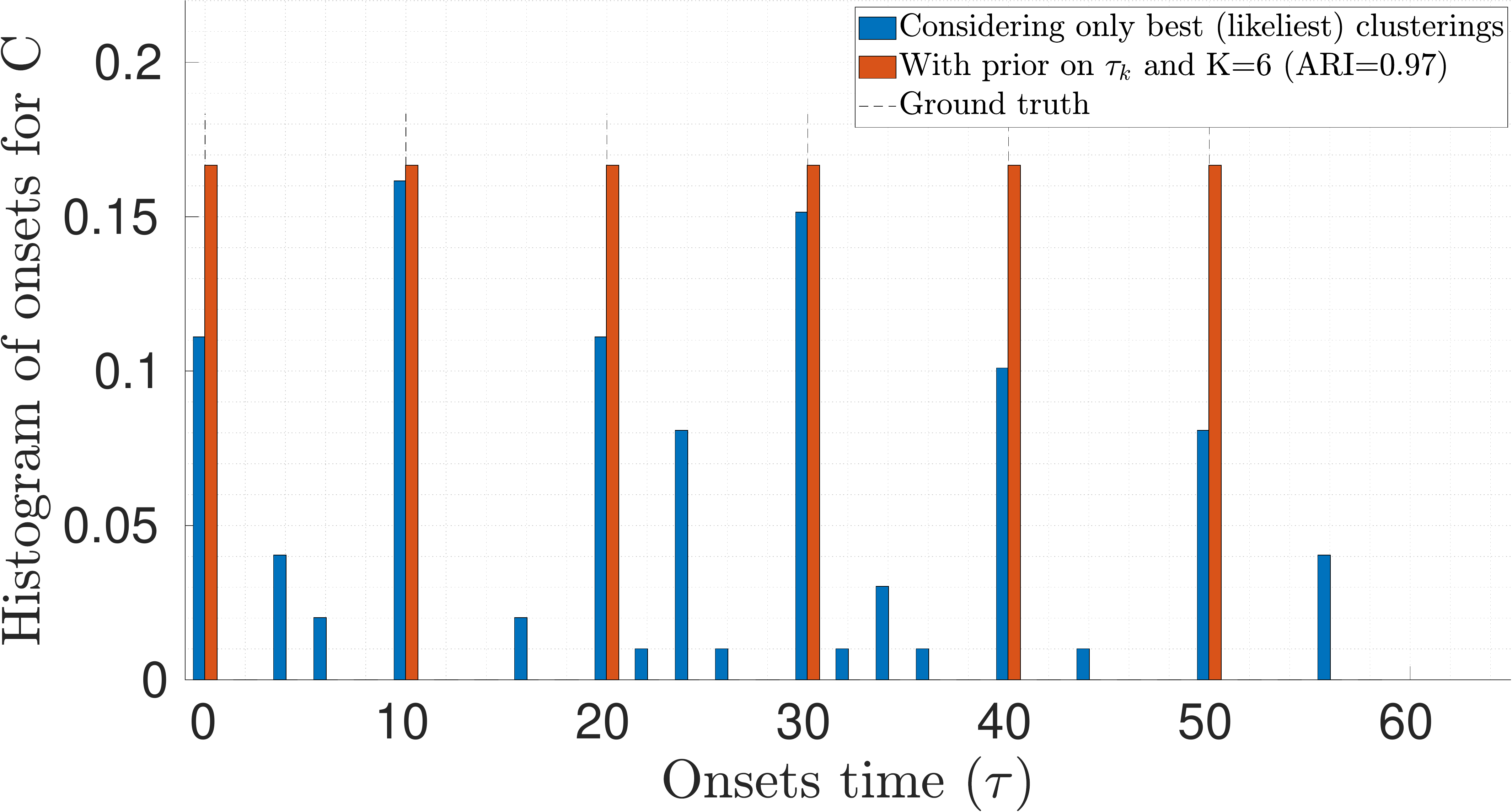}
\caption{Campaign C: (\textcolor{blue}{blue}) Histogram of $\tau_k$ estimates for the three initialization methods and $K$ ranging from 4 to 14, and (\textcolor{red}{red}) histogram with prior on onsets with regularization. \label{zefzegfergerg}}
\end{figure}

\begin{figure}
\centering
\includegraphics[width=0.75\textwidth]{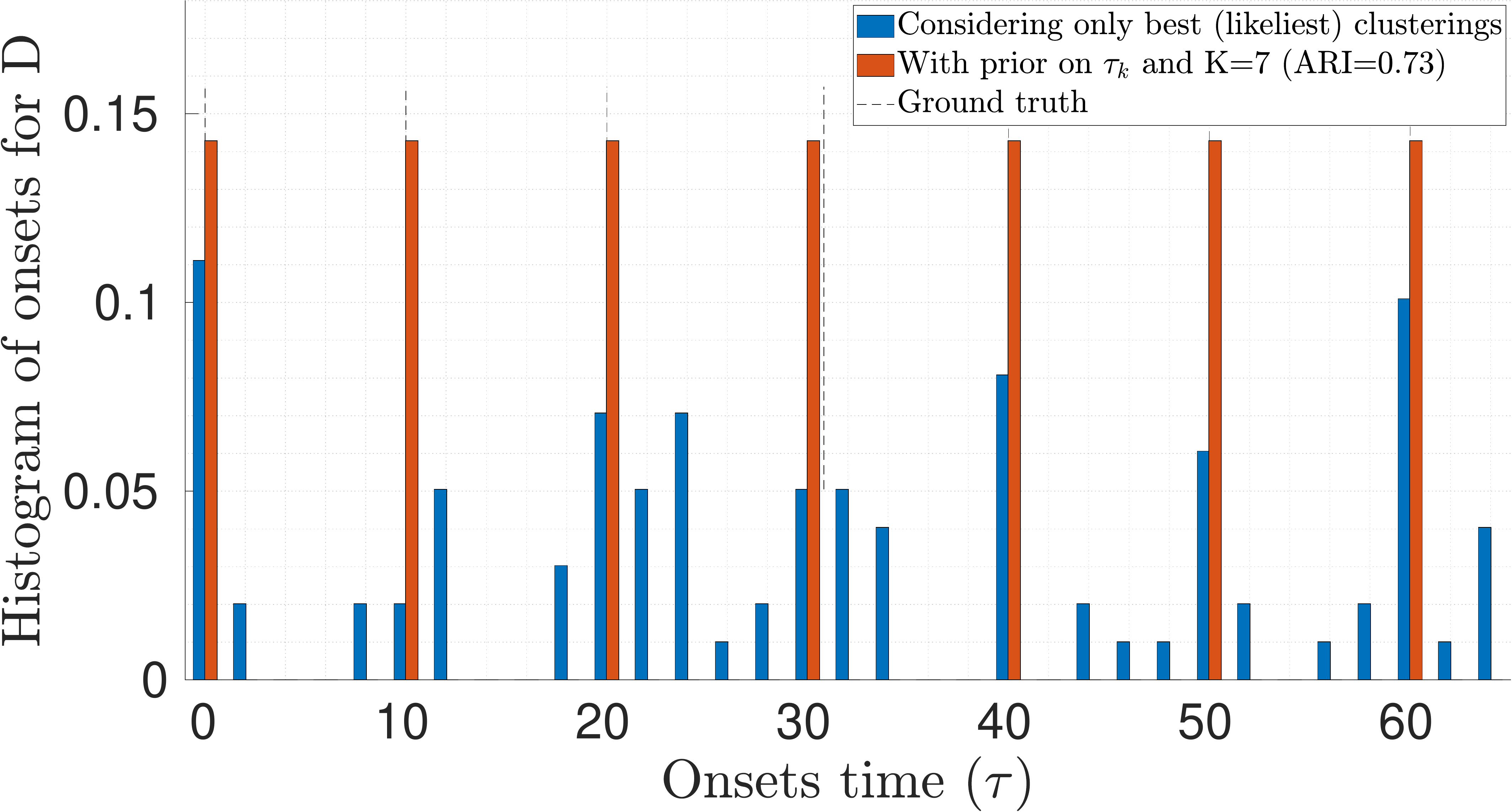}
\caption{Campaign D: (\textcolor{blue}{blue}) Histogram of $\tau_k$ estimates for the three initialization methods and $K$ ranging from 4 to 14, and (\textcolor{red}{red}) histogram with prior on onsets with regularization.\label{egregegegh}}
\end{figure}

\begin{figure}
\centering
\includegraphics[width=0.75\textwidth]{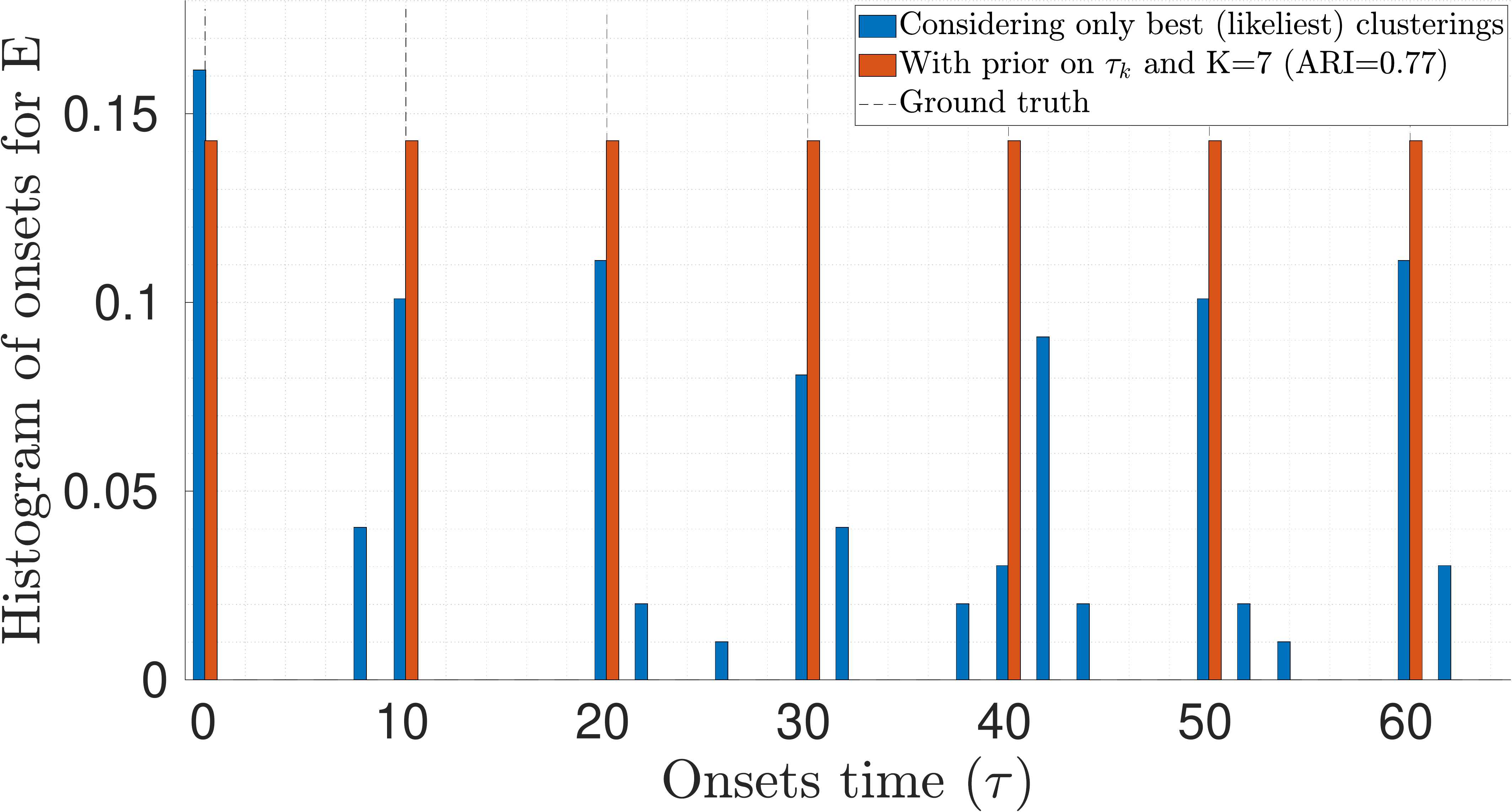}
\caption{Campaign E: (\textcolor{blue}{blue}) Histogram of $\tau_k$ estimates for the three initialization methods and $K$ ranging from 4 to 14, and (\textcolor{red}{red}) histogram with prior on onsets with regularization. \label{luiukiiyn}}
\end{figure}

\begin{figure}
\centering
\includegraphics[width=0.75\columnwidth]{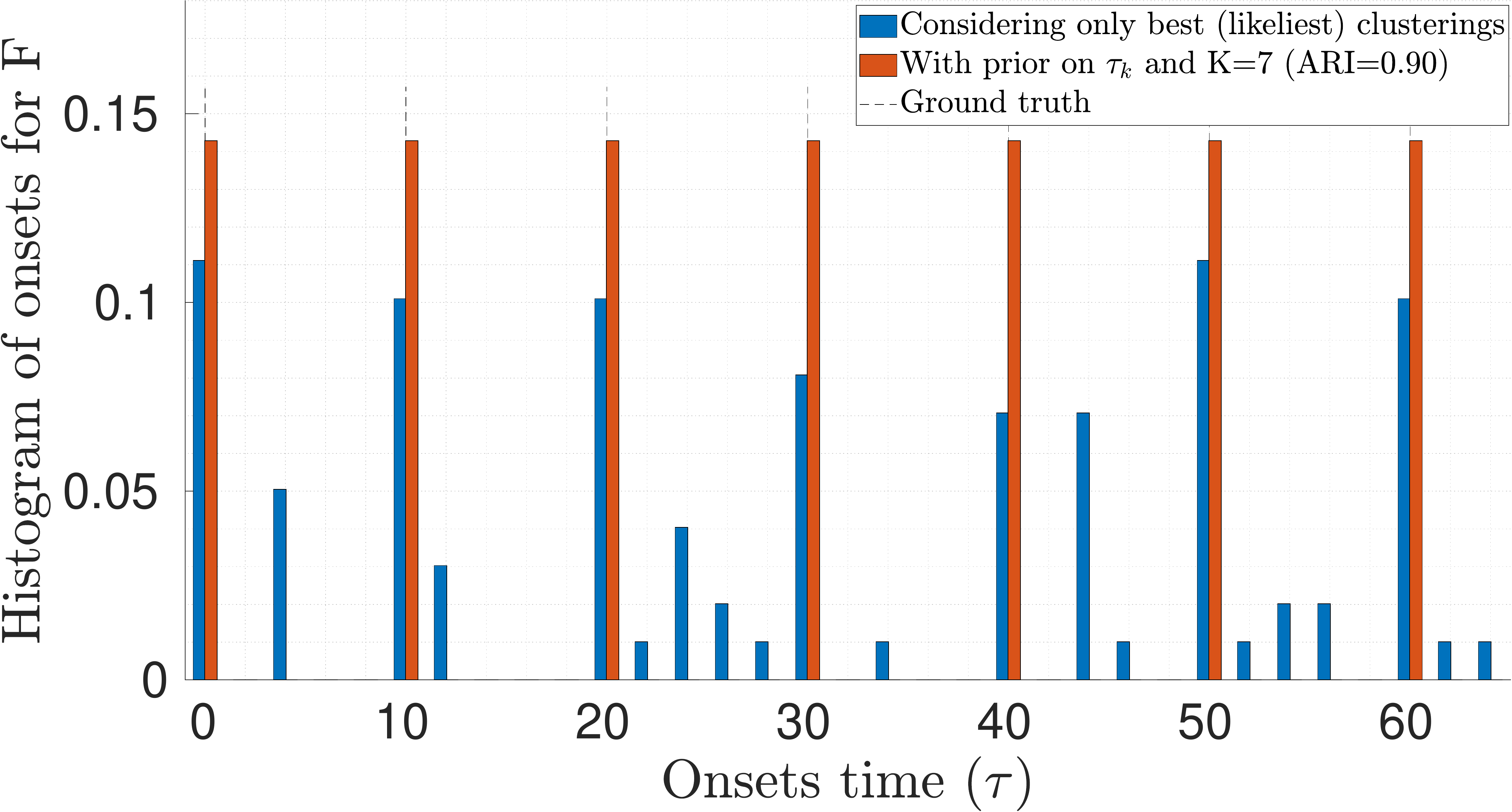}
\caption{Campaign F: (\textcolor{blue}{blue}) Histogram of $\tau_k$ estimates for the three initialization methods and $K$ ranging from 4 to 14, and (\textcolor{red}{red}) histogram with prior on onsets with regularization. \label{jtyhthrfke}}
\end{figure}

In Section \ref{intro}, we discussed  the importance of the onsets in the analysis of AE {{data set}} (represented by $\tau_k$ in \textsf{GMMSEQ}). These values were stacked for each campaign, independently of the type of initialization or the number of clusters. Figure \ref{ljklkjflzjelkf} shows how to interpret, in terms ot timeline,  Figures \ref{kqldkdkazlk} to \ref{jtyhthrfke} representing, in \textcolor{blue}{blue}, the normalised histograms of onsets estimated by \textsf{GMMSEQ} for each campaign. The dashed lines represent the instant when the tightening level was changed. These lines are separated by about 10 s (duration of each period) for each level; therefore, the horizontal axis, which represents time, can also be related to the tightening level: 0 s to 10 s corresponding to 60 cNm, 10 s to 20 s corresponding to 50 cNm, and so on until 60 to 70 s for 5 cNm (see Figure \ref{ljklkjflzjelkf}). The \textcolor{red}{red} bars represent the histograms when a prior on onsets is integrated through  \eqref{eq:Qbis} for K=7 clusters. 

\begin{table}[ht]
\centering
\caption{Performance of \textsf{GMMSEQ} with and without (``w/o'') prior on onsets, according to four 
measures of performance. Each measure is bounded in $[0,1]$ (the higher the better).}
{{
\begin{tabular}{|c||c|c|c|c|c|c|c|c|}
\hline
Campaign & \multicolumn{2}{|c|}{Precision} & \multicolumn{2}{|c|}{Recall} & \multicolumn{2}{|c|}{Entropy} & \multicolumn{2}{|c|}{ARI}  \\
   & with & w/o & with & w/o & with & w/o & with & w/o \\
\hline
B & 1.00 & 0.123 & 1.00 & 1.00 & 1.00 & 0.944 & 0.708 & 0.731 \\
\hline
C & 1.00 & 0.146 & 1.00 & 1.00 & 1.00 & 0.992 & 0.974 & 0.966 \\
\hline
D & 1.00 & 0.098 & 1.00 & 0.857 & 1.00 & 0.852 & 0.733 & 0.842 \\
\hline
E & 1.00 & 0.156 & 1.00 & 1.00 & 1.00 & 0.955 & 0.772 & 0.774 \\
\hline
F & 1.00 & 0.143 & 1.00 & 1.00 & 1.00 & 0.961 & 0.899 & 0.847 \\
\hline
\end{tabular}
}}
\end{table}

In Figures \ref{kqldkdkazlk} to \ref{jtyhthrfke}, a peak in the histogram means that several models provided similar values for $\tau_k$. Note that there are, for the \textcolor{blue}{blue} bars, $\sum_{k=4}^{14} k = 99$ estimates of the $\tau_k$ values, whereas there are $K=7$ values of $\tau_k$ for the \textcolor{red}{red} bars. In the latter case, we can observe that the prior on onsets allows us to obtain values of $\tau_k$ approximately equal to the ground truth (dashed lines) for all {{data sets}} and all tightening levels. The \textcolor{red}{red} bars depict a uniform distribution since the $\tau_k$ values are all different. 

For the purely unsupervised setting (in \textcolor{blue}{blue}), we can observe that the values of $\tau_k$ with the highest probability generally correspond to the instants {{when}} a change was made on the tightening level. This observation shows that \textsf{GMMSEQ} is able to discover the levels of tightening from the features. 
For each campaign we can make the following comments, remembering that the levels were approximately equal to 60 (cluster 1), 50 (cluster 2), 40 (cluster 3), 30 (cluster 4), 20 (cluster 5), 10 (cluster 6) and 5 cNm (cluster 7):

\begin{itemize}

\item Campaign B (Figure \ref{kqldkdkazlk}): Levels 60, 40, 30, 20 are precisely detected with a clear peak centered at the correct place. Levels 50, 10 and 5 shows less noticeable peaks but the bins in the histogram show modes that are well positioned around the expected positions. 

\item Campaign C (Figure \ref{zefzegfergerg}): For this campaign, a level is missing, which explains why there is no peak around 5 cNm (right-hand side);  the levels were indeed shifted by one level in the figure due to one missing level. Figure \ref{zefzegfergerg} shows clear peaks at the correct positions. There are also two additional peaks at 25s (30 cNm) and (5 cNm), which may be due to a change in the level during vibration tests. 

\item Campaigns D (Figure \ref{egregegegh}) and E (Figure \ref{luiukiiyn}): All levels are precisely detected with a clear peak centered at the correct location. 

\item Campaign F (Figure \ref{jtyhthrfke}): Levels 60, 50, 40, 20, 10, 5 are precisely detected with a clear peak (or two close peaks) centered at the correct location. For level 30 the peak is less noticeable. Some additional peaks appear around 5 s (middle of the period of 60 cNm) and 43 s (20 cNm), which may be due to a change in the level during vibration tests. 
\end{itemize}

From these figures, it can also be observed that \textsf{GMMSEQ} generated onsets with quite similar probability (except for campaign B). The bars in the histogram generally exhibit quite similar values for all tightening levels, which means that the number of AE signals for each tightening level are quite similar. Since the vibration has a fixed frequency independent of the tightening, the sources of AE signals seem to be activated in each cycle.


\begin{figure}[hbtp]
         \centering
         \includegraphics[width=0.8\textwidth]{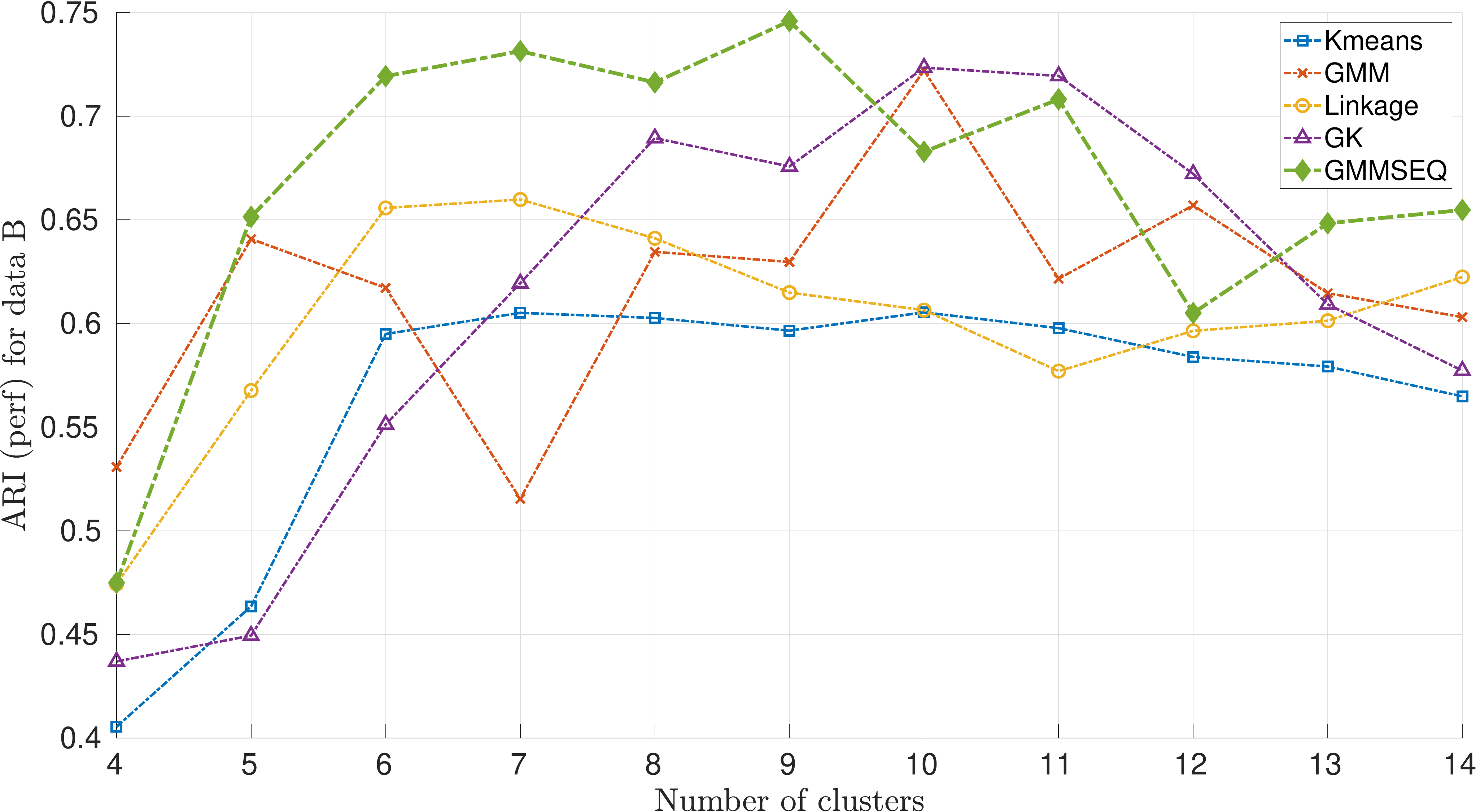}
        \caption{{Campaign B: Comparison between standard clustering methods and \textsf{GMMSEQ}}.\label{gfregergreg}}
\end{figure}

\begin{figure}[hbtp]
         \centering
         \includegraphics[width=0.8\textwidth]{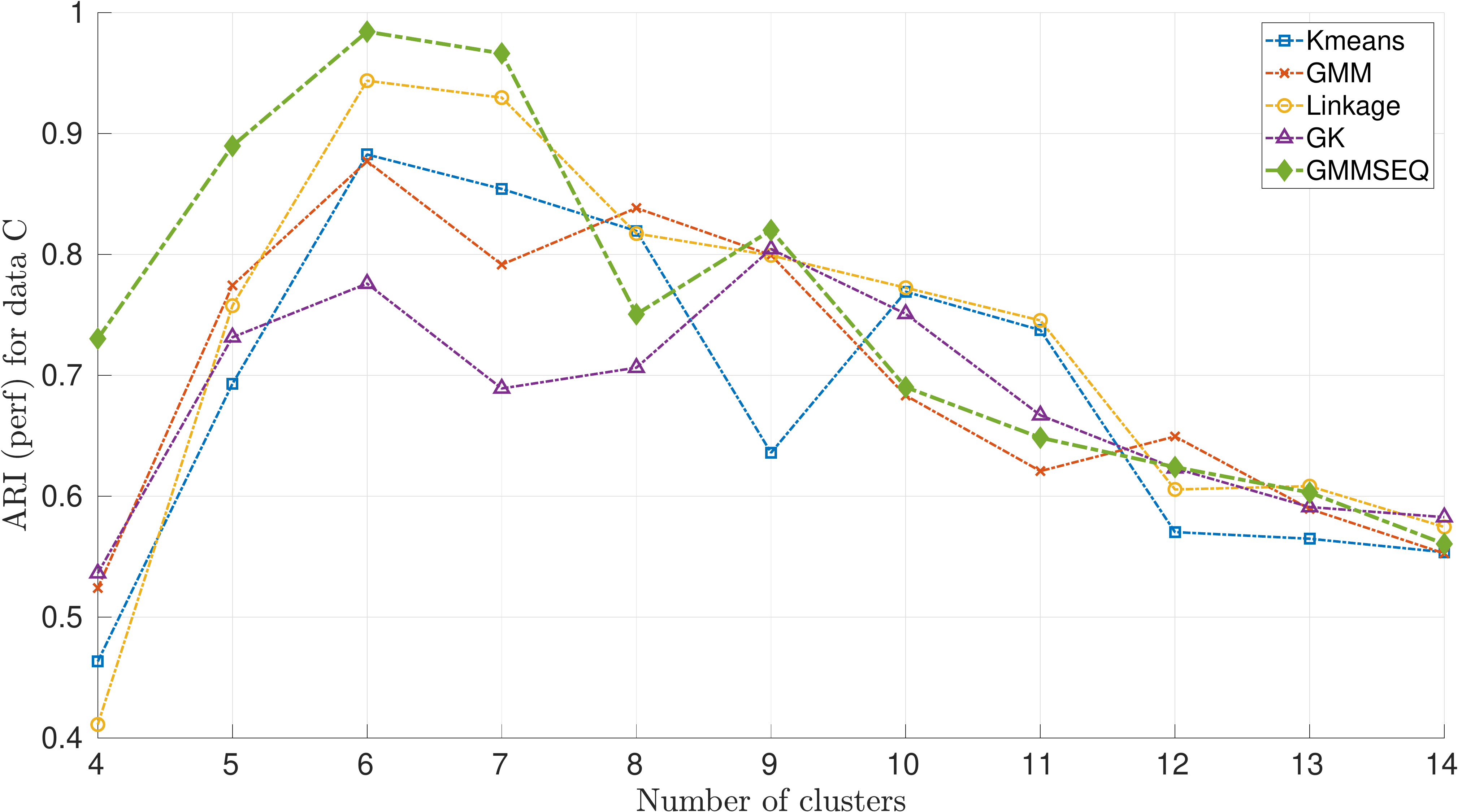}
        \caption{{Campaign C: Comparison between standard clustering methods and \textsf{GMMSEQ}.}\label{regegegeg}}
\end{figure}

\begin{figure}[hbtp]
         \centering
         \includegraphics[width=0.8\textwidth]{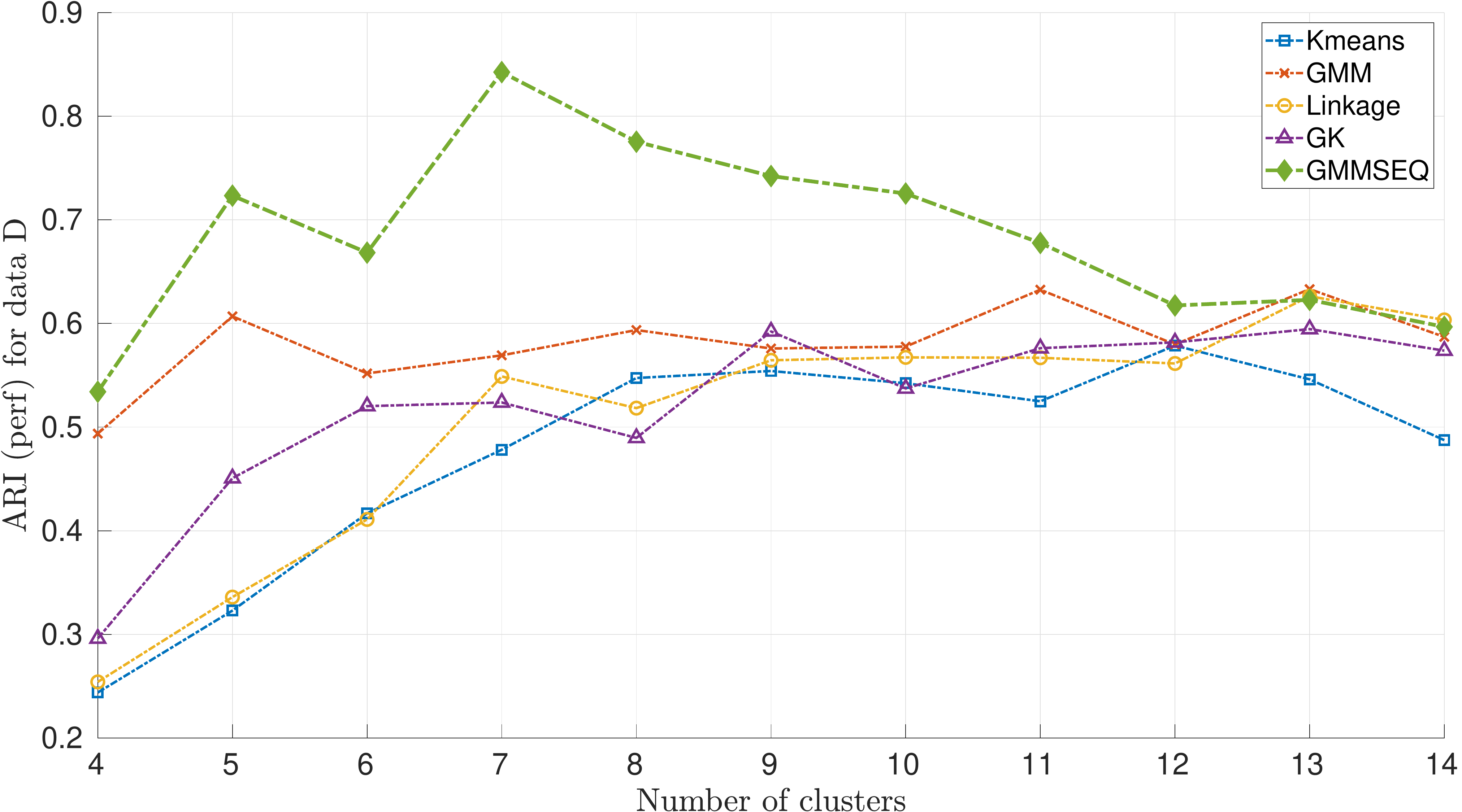}
        \caption{{Campaign D: Comparison between standard clustering methods and \textsf{GMMSEQ}.}\label{thtrjtrjrh}}
\end{figure}

\begin{figure}[hbtp]
     \centering
         \includegraphics[width=0.8\textwidth]{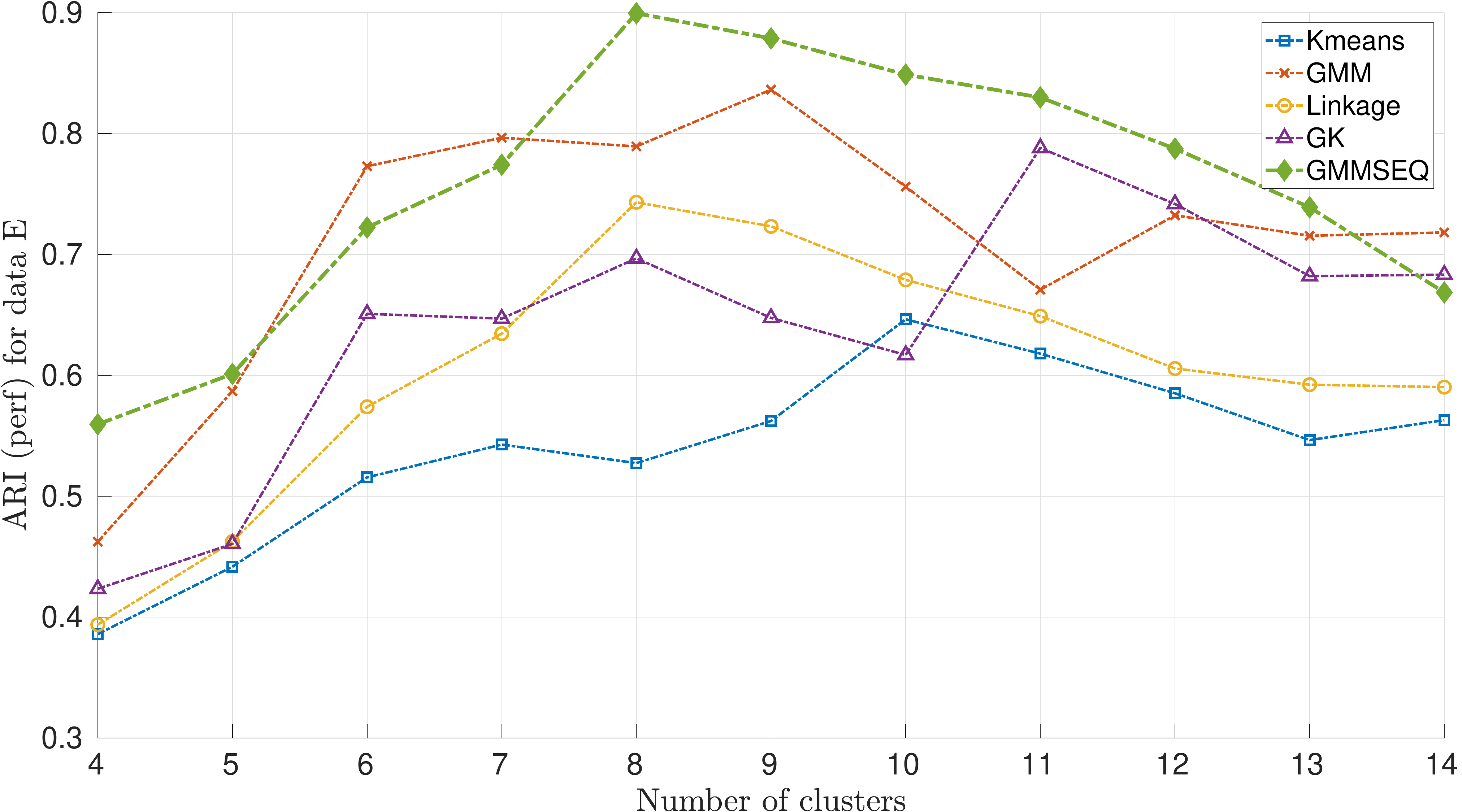}
        \caption{{Campaign E: Comparison between standard clustering methods and \textsf{GMMSEQ}.}\label{kjyukykytjrj}}
\end{figure}

\begin{figure}[hbtp]
     \centering
         \includegraphics[width=0.8\textwidth]{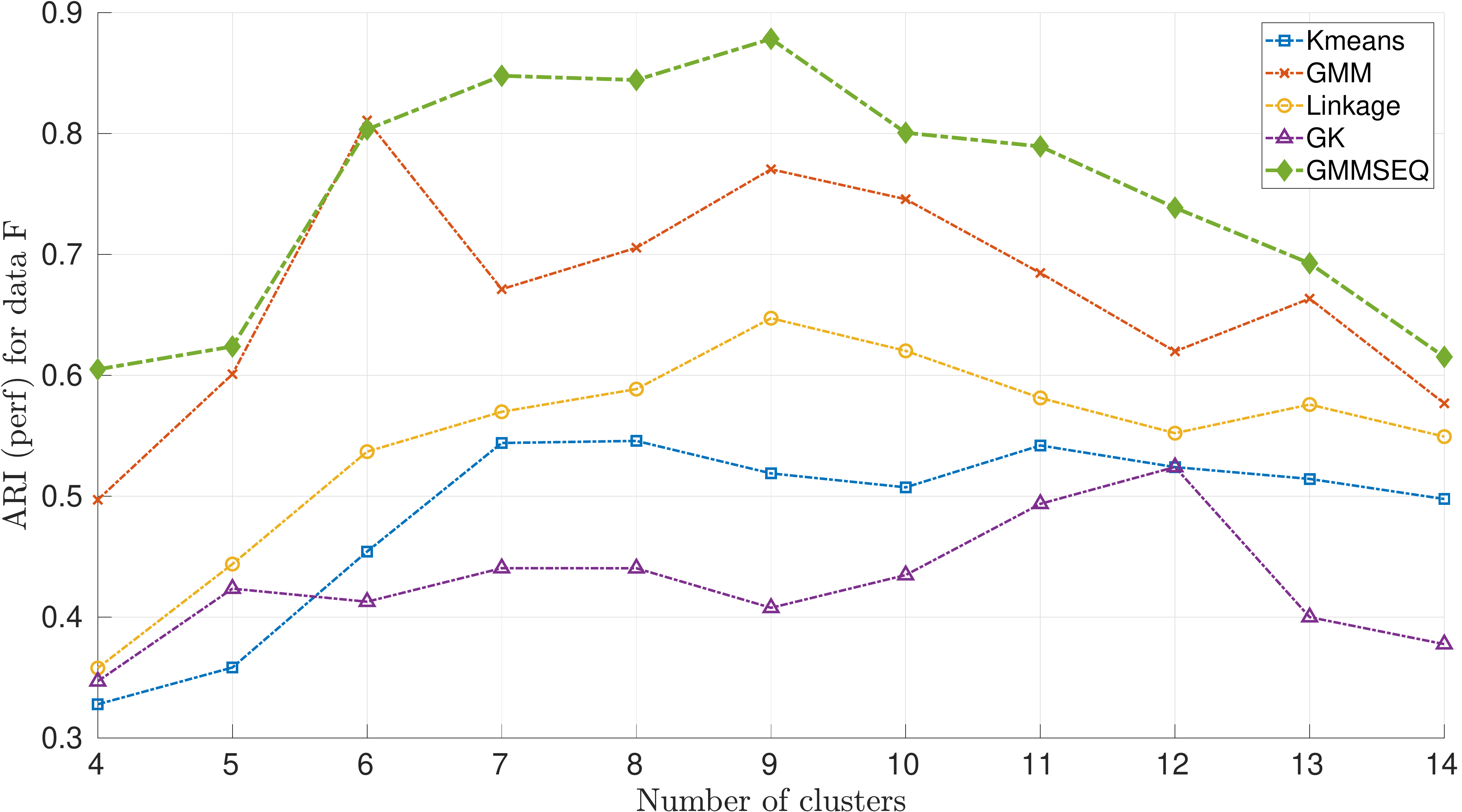}
        \caption{{Campaign F: Comparison between standard clustering methods and \textsf{GMMSEQ}.}\label{ouyjtghh}}
\end{figure}

{{
The evaluation of onsets was made by four criteria, each presenting a different view of the performance, namely precision, recall, entropy and ARI, detailed below. Each estimated onset was compared with the ground truth: If it remains within only $\pm 0.5$ second around a true value then it is considered as a correct estimation and qualified as a true positive (TP). If the onset is outside the interval, it is counted as a false positive (FP). If no onset is found in a given interval, a false negative (FN) is counted. TP, FN and FP are counted for all onsets. Then the accuracy is given by 
\[
Pr = \frac{TP}{TP+FP},
\]
which means that if a method has no false positive, the accuracy is $1$. A low accuracy corresponds to many FP. Therefore, a complementary criteria, called recall, is also used:
\[
Rec = \frac{TP}{TP+FN}
\]
which means that if a method does not miss any onset, the recall is $1$. A method with many FP can have a high recall if all expected onsets are found.  

We also expect  the methods to find onsets at similar locations when the number of clusters is changed (as explained in the analysis of histograms discussed above).  For that purpose, the normalized entropy of the cumulated onsets is computed as
\[
Ent = - \frac{ \sum_{k=1}^{6 \,or\, 7} p(k) \log_2 p(k) }{\log_2 K},
\]
where the sum is over the number of levels ($6$ for campaign C and $7$ for the others) and with 
\[
p(k) = \frac{\textrm{number of estimated onsets falling in } \pm 0.5 \textrm{ s around the truth}}{\textrm{total number of onsets estimated by the method}}
\]
which means that if a method provides $n$ onsets and there are $n/7$ onsets per level correctly located (or $n/6$ for campaign C), then the entropy is 1. An example of cumulated onsets with perfect location (leading to an entropy of $1$) is given with the red bars in Figures \ref{kqldkdkazlk} to \ref{jtyhthrfke}. 
}}

{{The fourth criterion is the Adjusted Rand Index (ARI) \cite{ari}. ARI is a corrected-for-chance (and, thus, more severe) version of the Rand Index used for clustering evaluation in the presence of a ground truth (in the present case, it corresponds to the tightening levels). The ARI is $1.00$ for perfect clustering and $0$ in case of random assignment of clusters or totally wrong clustering. The results are summarized in Tables \ref{tabB}, \ref{tabC}, \ref{tabD}, \ref{tabE} and \ref{tabF} for all campaigns (B to F respectively), all criteria (precision, recall, entropy and ARI) and several clustering methods (results will be described below).
}}

For comparison purposes, four standard clustering methods were applied: K-means, a standard GMM, a hierarchical clustering (HC) using the Ward method {{and the Gustafson-Kessel (GK) algorithm}} (implementations are available in the MATLAB Statistics and Machine Learning (SML) toolbox and on MATLAB Central). The same matrix of features was used for all methods, including \textsf{GMMSEQ}. The K-means algorithm was run 10 times and the model with the smallest sum of squared-distance was selected. GMM was run 10 times with full covariance matrices for each cluster and the model with the highest likelihood was selected. {{A GMM was used to initialize the Gustafson-Kessel algorithm  with fuzziness degree set to $1.5$ as proposed in \cite{ramasso2015unsupervised}. For these three methods, the onsets are defined by the time of the first occurrence of every cluster. For \textsf{GMMSEQ}, onsets were given by $\boldsymbol{\tau}$ values.}}

{{Figures \ref{gfregergreg} to \ref{ouyjtghh} depict the ARI for all methods and all campaigns with respect to the number of clusters. In each figure, the curves with diamond markers represent the ARI for \textsf{GMMSEQ}. For this method, it can be observed that the maximum is generally observed for 6 to 9 clusters which is consistent with the expected number of tightening levels. These figures show that \textsf{GMMSEQ} globally outperforms K-means, GK, HC and GMM clustering methods. A more detailed analysis is provided in Tables \ref{tabB} to \ref{tabF} using the four aforementioned criteria. The ARI is the average over 6 to 9 clusters (and 5-8 for campaign C). In the tables, the ``\textbf{bold}'' font is used for the best overall for a given criterion, while an ``\underline{underlining}'' is used when comparing only \textsf{GMMSEQ} and the original GMM. Therefore, if one of these two methods is both underlined and bold then it represents the best of all methods for a given criterion.

The tables show that:
\begin{itemize}
\item For all campaigns and all methods, the precision is very low meaning that no method is able to detect all onsets at $\pm 0.5$ second around the true value.  
\item For the recall criterion, \textsf{GMMSEQ} provides a value of $1.0$ for four campaigns, while providing a value of $0.85$ (against 0.714 only for GMM) for campaign D. It globally outperforms all methods. The recall of the original GMM is never better than the recall of \textsf{GMMSEQ}. 
\item According to the entropy, \textsf{GMMSEQ} provides the best performance for four campaigns, and is only outperformed by HC for campaign D ($0.852$ against $0.921$). The entropy of the original GMM is never better than the entropy of \textsf{GMMSEQ}. 
\item Finally, \textsf{GMMSEQ} provides the best ARI for all campaigns. Compared to the GMM, we have +13.3\%, +13.3\%, +19\%, +1.3\% and +10.3\% for, respectively, campaign B, C, D, E and F. The ARI of the original GMM is thus never better than the ARI of \textsf{GMMSEQ}. 
\end{itemize}
These tables show that there is no best method for all campaigns and according to all criteria. However, from a quantitative point of view, \textsf{GMMSEQ} globally outperforms the GMM which shows the significance of the  proposed model. 
}}

\begin{table}[htpb]
\centering
{{
\caption{{Campaign B: Performance}\label{tabB}}
\begin{tabular}{|l||l|l|l|l|}
\hline 
Algo/perf & precision & recall & entropy & ARI \\ 
\hline  \hline
Kmeans & 0.066 & 0.571  & 0.597 & 0.601 \\ 
\hline 
GMM & 0.100 & 0.857  & 0.805  & 0.589\\ 
\hline 
HC & 0.067 & 0.571  & 0.712 & 0.652 \\ 
\hline 
GK & 0.081 & 0.714  & 0.706  & 0.620\\ 
\hline 
GMMSEQ & \underline{\textbf{0.123}} & \underline{\textbf{1.00}}  & \underline{\textbf{0.944}} & \underline{\textbf{0.722}} \\ 
\hline 
\end{tabular}
}}
\end{table}

\begin{table}[htpb]
\centering
{{
\caption{{Campaign C: Performance}\label{tabC}}
\begin{tabular}{|l||l|l|l|l|}
\hline 
Algo/perf & precision & recall & entropy & ARI \\ 
\hline  \hline
Kmeans & 0.188 & \textbf{1.00}  & 0.853 & 0.810 \\ 
\hline 
GMM & \underline{\textbf{0.200}} & \underline{\textbf{1.00}}  & 0.931 & 0.814 \\ 
\hline 
HC & 0.118 & 0.667  & 0.739 & 0.877\\ 
\hline 
GK & 0.177 & \textbf{1.00}  & 0.963 & 0.732 \\ 
\hline 
GMMSEQ & 0.146 & \underline{\textbf{1.00}}  & \underline{\textbf{0.992}} & \underline{\textbf{0.947}} \\ 
\hline 
\end{tabular} 
}}
\end{table}

\begin{table}[htpb]
\centering
{{
\caption{{Campaign D: Performance}\label{tabD}}
\begin{tabular}{|l||l|l|l|l|}
\hline 
Algo/perf & precision & recall  & entropy & ARI \\ 
\hline  \hline
Kmeans & 0.152 & \textbf{1.00}  & 0.799 & 0.480  \\ 
\hline 
GMM & 0.086 & 0.714  & 0.791 & 0.572 \\ 
\hline 
HC & \textbf{0.212} & \textbf{1.00}  & \textbf{0.921} & 0.493\\ 
\hline 
GK & 0.098 & 0.714  & 0.801  & 0.511\\ 
\hline 
GMMSEQ & \underline{0.097} & \underline{0.857}  & \underline{0.852} & \underline{\textbf{0.762}}\\ 
\hline 
\end{tabular}
}}
\end{table}

\begin{table}[htpb]
\centering
{{
\caption{{Campaign E: Performance}\label{tabE}}
\begin{tabular}{|l||l|l|l|l|}
\hline 
Algo/perf & precision & recall  & entropy & ARI\\ 
\hline  \hline
Kmeans & 0.065 & 0.429  & 0.561 & 0.529 \\ 
\hline 
GMM & 0.146 &  \underline{\textbf{1.00}}  & 0.891 & 0.786\\ 
\hline 
HC & 0.069 & 0.571  & 0.600 & 0.651 \\ 
\hline 
GK & 0.058 & 0.429  & 0.547  & 0.665\\ 
\hline 
GMMSEQ & \underline{\textbf{0.156}} & \underline{\textbf{1.00}} & \underline{\textbf{0.956}} & \underline{\textbf{0.799}} \\ 
\hline 
\end{tabular}
}}
\end{table}

\begin{table}[htpb]
\centering
{{
\caption{{Campaign F: Performance}\label{tabF}}
\begin{tabular}{|l||l|l|l|l|}
\hline 
Algo/perf & precision & recall & entropy  & ARI\\ 
\hline  \hline
Kmeans & 0.062 & 0.571  & 0.646 & 0.515\\ 
\hline 
GMM & \underline{\textbf{0.146}} & \underline{\textbf{1.00}} & 0.896 & 0.729\\ 
\hline 
HC & 0.078 & 0.714  & 0.774  & 0.565\\ 
\hline 
GK & 0.100 & 0.714  & 0.728  & 0.431\\ 
\hline 
GMMSEQ & 0.143 & \underline{\textbf{1.00}}  & \underline{\textbf{0.961}}  & \underline{\textbf{0.832}}\\ 
\hline 
\end{tabular}
}}
\end{table}

\begin{figure}[htpb]
         \centering
         \includegraphics[width=0.7\textwidth]{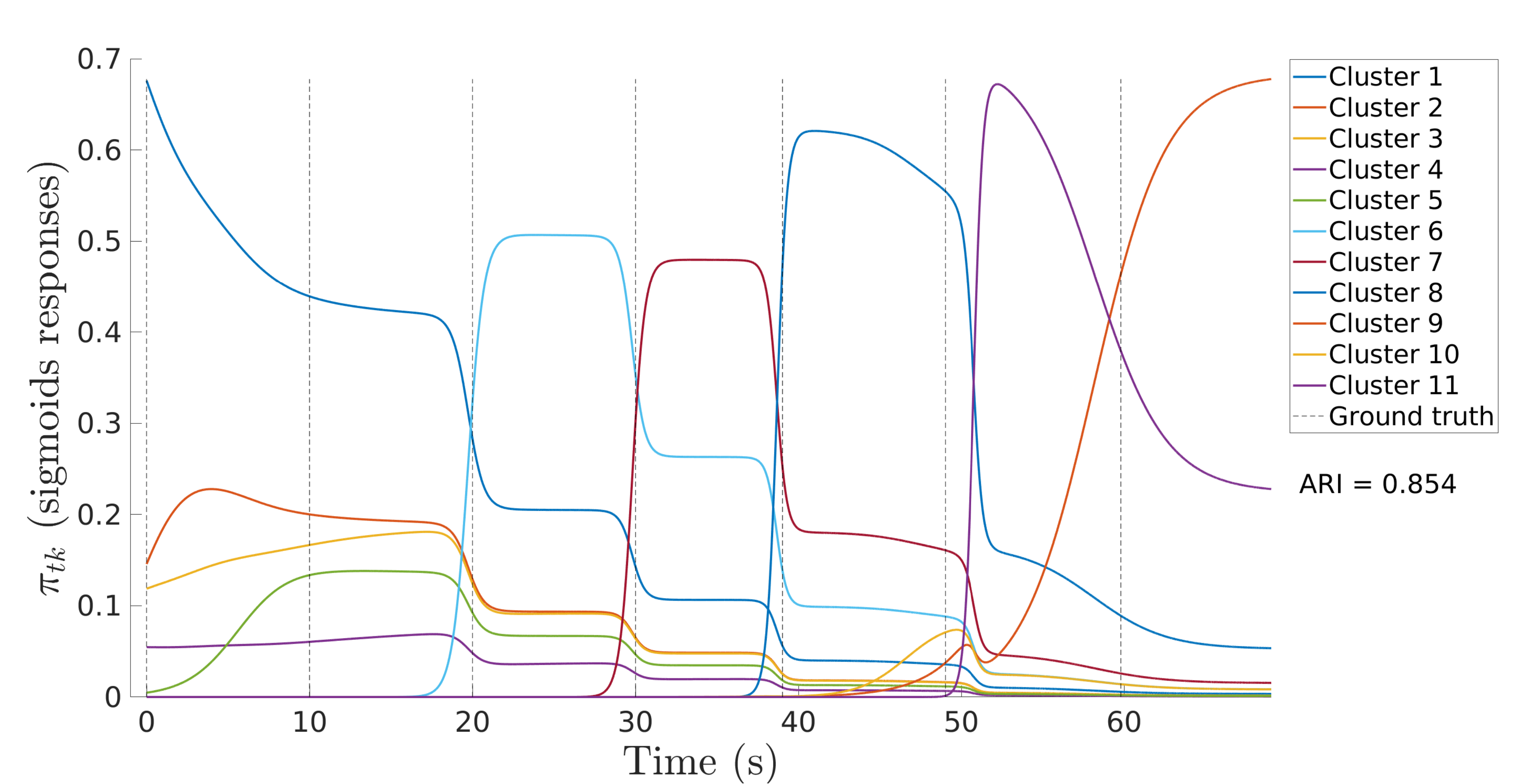}
        \caption{{Campaign B: $\pi_{tk}$ values provided by \textsf{GMMSEQ}}.\label{gfregergreg22}}
\end{figure}

\begin{figure}[htpb]
         \centering
         \includegraphics[width=0.7\textwidth]{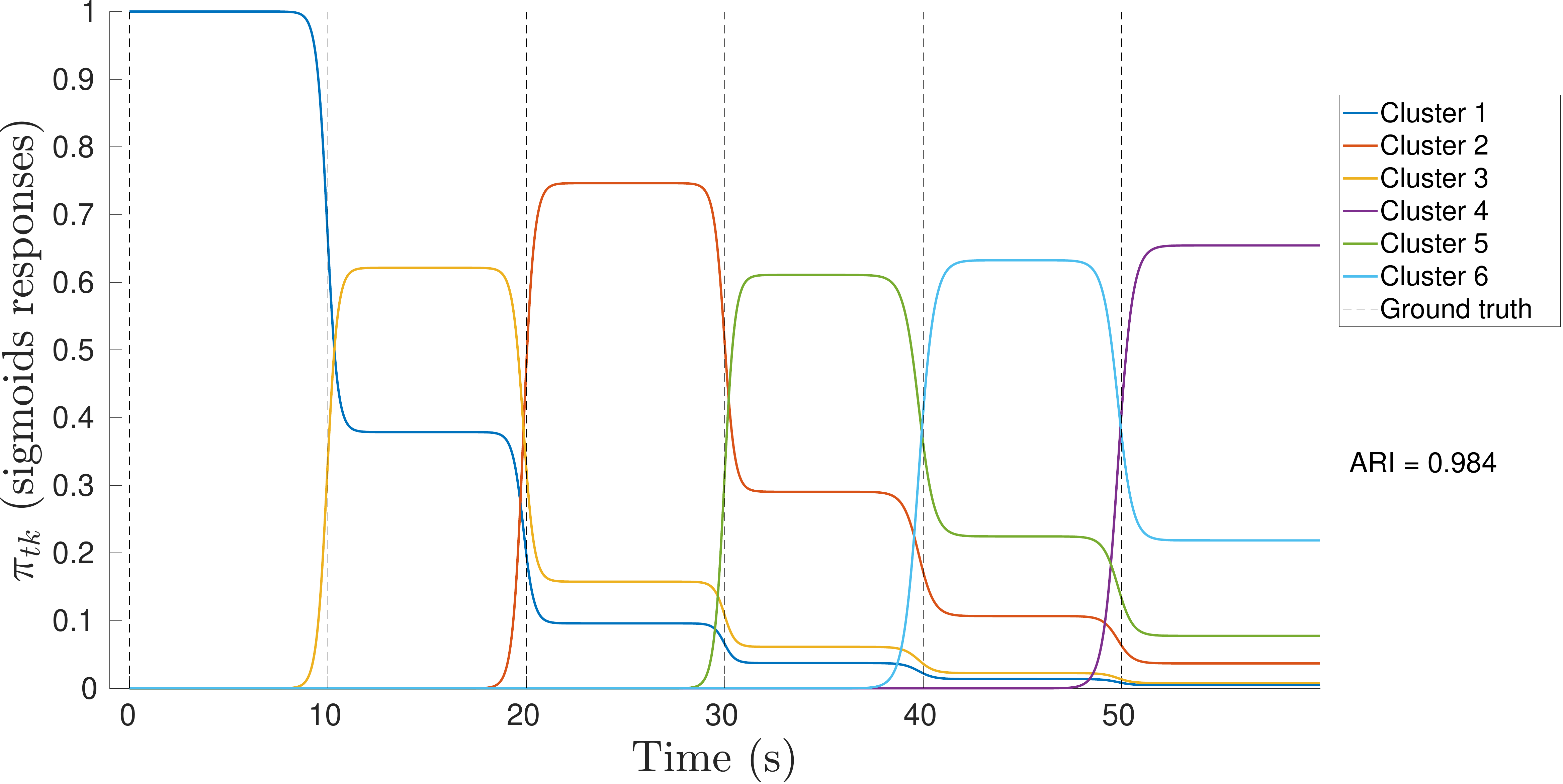}
        \caption{{Campaign C: $\pi_{tk}$ values provided by \textsf{GMMSEQ}}.\label{gfregergreg23}}
\end{figure}

\begin{figure}[htpb]
         \centering
         \includegraphics[width=0.7\textwidth]{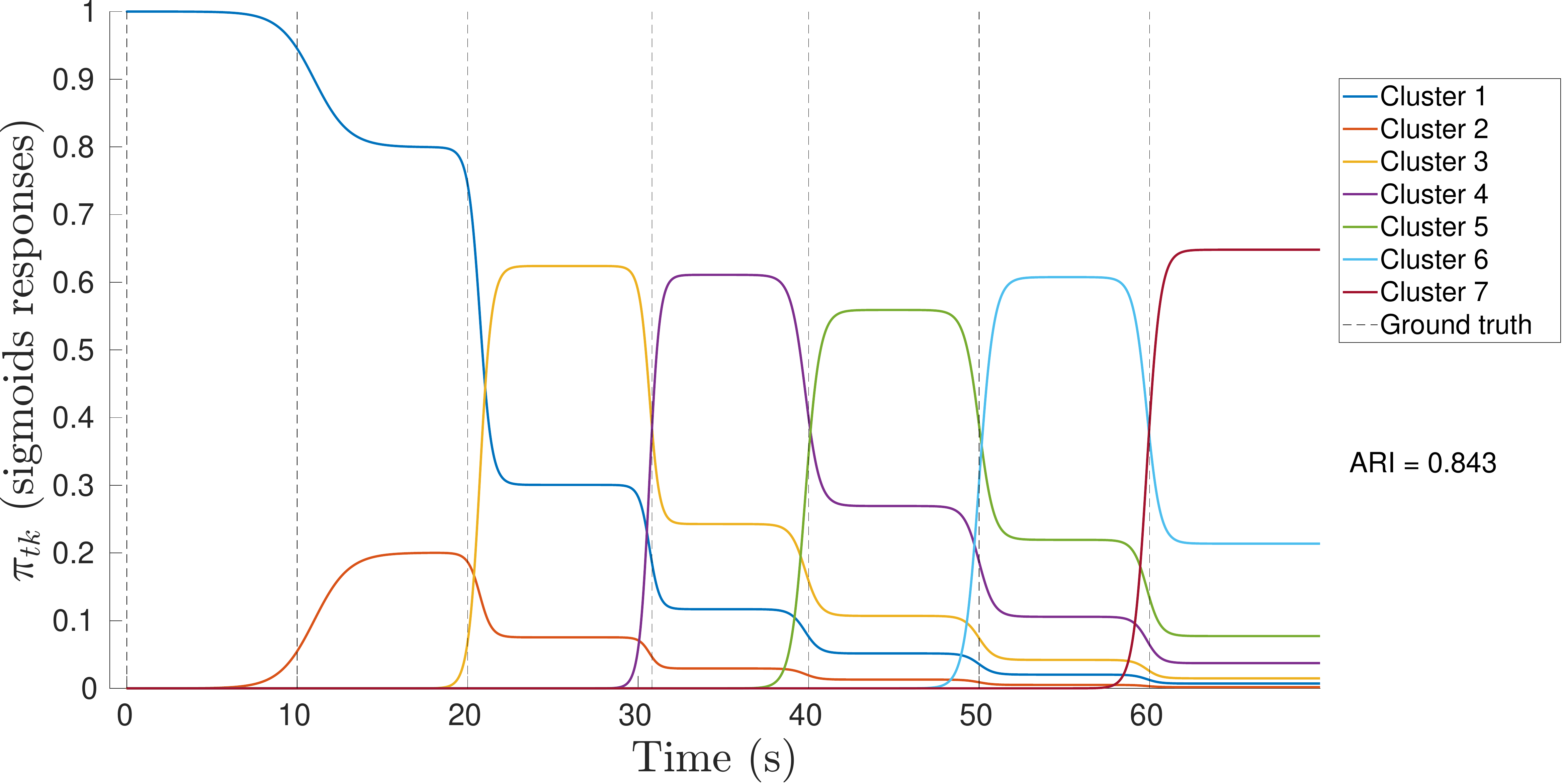}
        \caption{{Campaign D: $\pi_{tk}$ values provided by \textsf{GMMSEQ}}.\label{gfregergreg24}}
\end{figure}

\begin{figure}[htpb]
         \centering
         \includegraphics[width=0.7\textwidth]{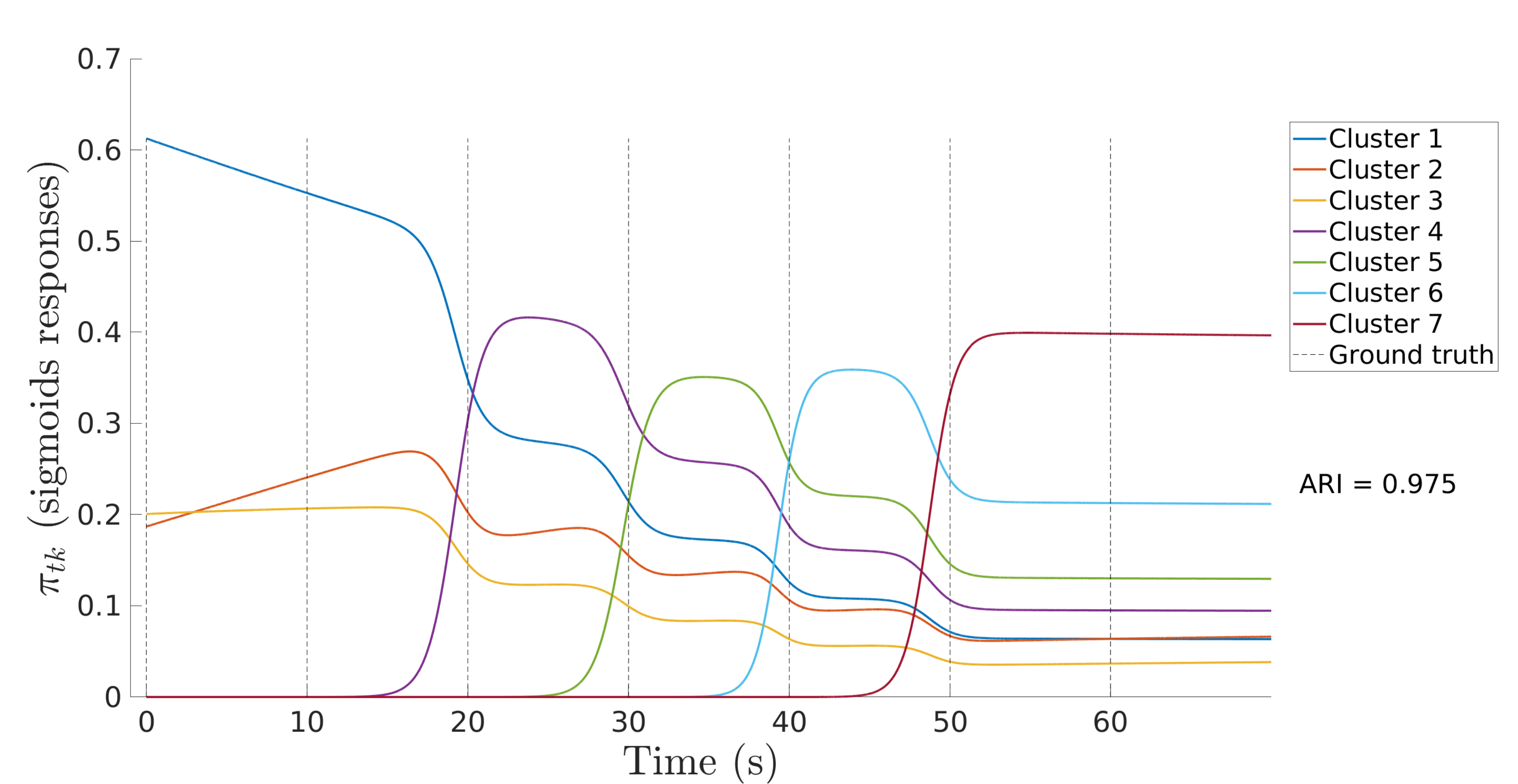}
        \caption{{Campaign E: $\pi_{tk}$ values provided by \textsf{GMMSEQ}}.\label{gfregergreg25}}
\end{figure}

\begin{figure}[htpb]
         \centering
         \includegraphics[width=0.7\textwidth]{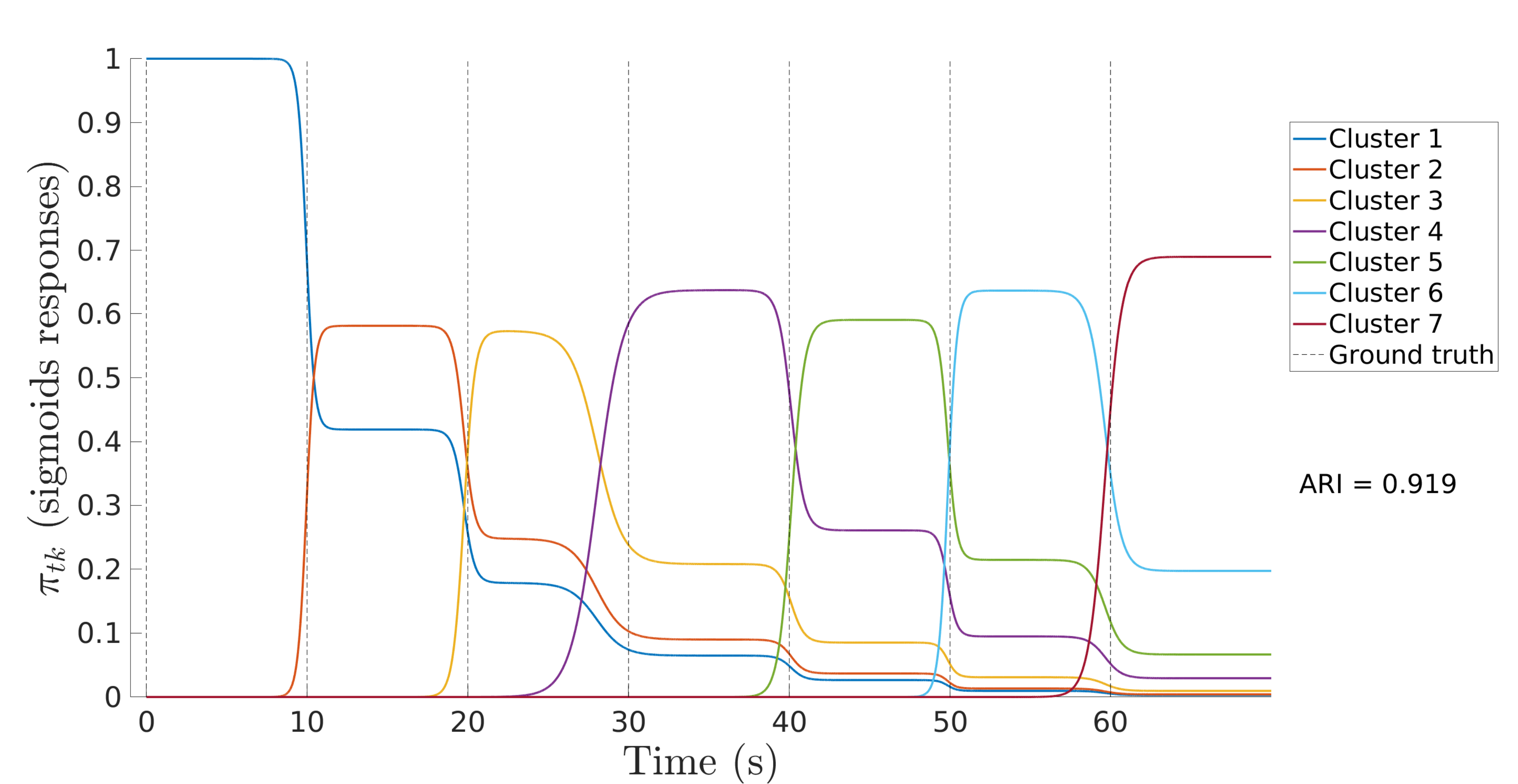}
        \caption{{Campaign F: $\pi_{tk}$ values provided by \textsf{GMMSEQ}}.\label{gfregergreg26}}
\end{figure}

{{
The previous results and analyses demonstrated the usefulness of considering onsets, $\boldsymbol{\tau}$, as parameters to be identified from AE data together with clusters parameters. 
The simulated data also allowed us to show that our model was able to perfectly recover not only the onsets but also the levels of activation, $\boldsymbol{\beta}$, and kinetics, $\boldsymbol{\gamma}$ for four clusters with different behavior. In order to conclude this study, we propose an analysis of these parameters, at the core of \textsf{GMMSEQ}, on the real data. 

Figures \ref{gfregergreg22} to \ref{gfregergreg26} illustrate the degrees of activation, $\pi_{tk}$, in each campaign computed from the \textsf{GMMSEQ} model with the highest ARI. The value $\pi_{tk}$ is obtained from  $\boldsymbol{\tau}$,  $\boldsymbol{\beta}$ and  $\boldsymbol{\gamma}$ using Eq.~\ref{eq:props}. Onsets, kinetics and levels of activation are well illustrated on these figures. It is expected that the levels of activation (the height of curves) should be similar for all tightening levels since we have about the same number of cycles per tightening level. This is well illustrated on the figures. Ideally, level 1 should tend to 1 due to the normalization (Eq.~\ref{eq:props}), which is the case for campaigns C (Fig. \ref{gfregergreg23}, D (Fig. \ref{gfregergreg24}) and F (Fig. \ref{gfregergreg26}). For campaign B and E, three or more clusters coexist at the start of the test (60 cNm) but the highest probability is still assigned to the correct cluster. 

Another expectation is that the kinetics should be similar between levels and the slope of the changes in the values $\pi_{tk}$ should be quite steep because the tightening levels are modified abruptly according to \cite{ORIONdata}. This is also well illustrated on the figures where the slopes are globally similar and steep for each new onset detected. It is worth noting that gradual onsets are obtained for campaign E characterized by lower $\boldsymbol{\gamma}$ values (estimated as $[0,    3.8587,    2.9722,    1.2202,    2.4498,    3.4912,    1.9343]$ seconds so an average of $0.719 \pm 0.652$ s at one standard deviation) than for the other campaigns (with average of $1.298 \pm 1.31$ s for B, $2.77 \pm 1.46$ s for C, $2.07 \pm 1.17$ s for D and $2.28 \pm 1.35$ s for F). As seen on Figure \ref{gfregergreg23}, the slopes for C are the steepest ones and the $\boldsymbol{\gamma}$ values are also the largest as expected. The differences between values are sufficiently important to strongly modify the slopes through the exponential in the sigmoid function (Eq. \ref{eq:alpha}).

The onsets are not always sufficient to characterize the performance. Indeed, despite two confusions illustrated for campaign E (levels 1/2 and 6/7), the clustering is close to be perfect with an ARI around 0.975. It means that the mixture parameters (means and covariances of \textsf{GMMSEQ}) also play a key role in the performance (as expected since these parameters pave the feature space). 

Despite high ARI values compared to Tables \ref{tabB}-\ref{tabF} (in these tables, the best model was selected based on maximum likelihood, not based on the ARI), the location of the sigmoids is not perfect: For example, for campaign D, we can observe that the second level (50 cNm) is not detected properly with a confusion with level 1 (60 cNm). Level 2 is detected but its probability is around 0.2 against 0.8 for level 1. With these curves, the end-user is thus able to visually identify some confusions which is not possible with other methods. A similar behavior can be observed for campaign B. For this campaign, the best ARI is obtained with 11 clusters with several of them concentrated around levels 1 and 2. These confusions partly explain the result of 0.762 (resp. 0.722) for the ARI of campaign D in Table \ref{tabD} (resp. campaign B in Table \ref{tabB}). It is worth noting that, in both cases, the main confusion arises between two consecutive levels (60 cNm and 50 cNm) which is more understandable than if the confusion were between very distinct levels. 
}}

\section{Conclusion}

\textsf{GMMSEQ} {{is a new clustering method introduced to manage continuous timestamps attached to AE signals. 
Timestamps are exploited during the clustering process allowing one to gain new insights into the AE data streams. In addition, cluster onsets, growth rates and  levels of activation through time are estimated together with the parameters of their distributions in the feature space.}} {{Therefore, this new clustering method has unique characteristics that stand out from existing methods for AE data interpretation. 

\textsf{GMMSEQ} represents the first clustering method specifically developed for AE data, and it is the only method that allows AE users to estimate, directly from data, parameters related onsets, growth and kinetics. To our knowledge, there is no method with similar features in the literature. 

The performance of the method has been demonstrated on simulated and real {{data sets}}. Close to perfect results were obtained on simulated data with noise. On real data, we demonstrated the relevance of the clusters during loosening of bolted joints. The comparison with four standard clustering methods and according to different performance criteria shows that \textsf{GMMSEQ} not only provides useful qualitative indications about the timeline of clusters, but also has better performance in terms of cluster characterization. }}

{{This work opens up several perspectives. First, the approach can be extended to other mixture models considering non-Gaussian distributions. Our shared code includes, for example, an extension to a mixture of multivariate Student-t distributions, which is not described in this paper.  

The possibility to include prior information about the cluster onsets  was addressed in the paper using a regularization of the objective function. Prior information about some other parameters such as the cluster centers could be exploited too, which could potentially improve the convergence of the algorithm to ``better'' estimates with a clear physical interpretation. The possibility of incorporating physical knowledge  in the objective function could also be investigated. 

Another direction to explore concerns the optimization procedure. In this paper, we have assumed that all data are available at once (offline or batch analysis).  The extension to online clustering with evolving parameters is a promising perspective for application, for example, to structural health monitoring or statistical process control.

Finally, while being developed with the goal to process AE data, this clustering method can be applied to other data sets containing data streams with gradually emerging clusters. 
 }}

\section*{Acknowledgement}

This work was partly carried out in the framework of the EIPHI Graduate school (contract ANR-17-EURE-0002) and the project RESEM-COALESCENCE funded by the Institut de Recherche Technologique Mat\'eriaux M\'etallurgie Proc\'ed\'es (IRT M2P) and Agence Nationale de la Recherche (ANR). The authors are also thankful to MIFHySTO and AMETISTE platforms. Finally, the authors thank reviewers for their effort to review the manuscript which helped us in greatly improving its quality.

\section*{Codes and data}

Codes to reproduce all results on clustering are available on Github at \url{https://github.com/emmanuelramasso/MIXMOD_SEQUENTIAL}. Data are also available in their raw format at \url{http://dx.doi.org/10.7910/DVN/FBRDU0}. The features used as inputs for \textsf{GMMSEQ} can finally be downloaded at \url{https://drive.google.com/drive/folders/1H413RxYu4ya7YMEgF_lTh_fHr7flvvOO?usp=sharing}.

\appendix

\section{Gradient of $Q$ with respect to $g_k$, $b_k$ and $\xi_k$}
\label{mzjkdpzakodoakda}

The gradients of the auxiliary function $Q$ with respect to the variables of interest are given below. We first start by expressing the derivatives with respect to the instrumental variables, namely $g_k,b_k,\xi_k$ for each component in the mixture:
\begin{subequations}
\begin{align}
\label{eq:dQdg}
\parpapb{Q}{}{g}{k} &=  \parpapb{Q}{}{\gamma}{k} \parpapb{\gamma}{k}{g}{k}= 2 g_k \parpapb{Q}{}{\gamma}{k},  \quad k=2,\ldots,K\\
\label{eq:dQdb}
\parpapb{Q}{}{b}{k} &= \parpapb{Q}{}{\beta}{k}  \parpapb{\beta}{k}{b}{k}= 2 b_k \parpapb{Q}{}{\beta}{k},  \quad k=2,\ldots,K\\
\label{eq:dQdxi}
\parpapb{Q}{}{\xi}{k} &=  \parpapb{Q}{}{\tau}{k} \parpapb{\tau}{k}{\xi}{k}= \parpapb{Q}{}{\tau}{k} \tau_k \left(1-\frac{\tau_k}{T}\right),  \quad k=2,\ldots,K,
\end{align}
\label{gradALL}
\end{subequations}
where \eqref{eq:dQdxi} uses the following property of the logistic function $\Lambda(u)=1/(1-\exp(-u))$:  $\Lambda'(u)=\Lambda(u)[1-\Lambda(u)]$. The derivatives with respect to $\beta_k$, $\gamma_k$ and $\tau_k$ are given by
\begin{subequations}
\begin{align}
\label{eq:dQdbeta}
\parpapb{Q}{}{\beta}{k} &=  \sum_{l=1}^K \sum_{i=1}^N \parpapb{Q}{}{\pi}{il} \parpapb{\pi}{il}{\alpha}{ik} \parpapb{\alpha}{ik}{\beta}{k},  \quad k=2,\ldots,K, \quad i=1,\ldots,N\\
\label{eq:dQdgamma}
\parpapb{Q}{}{\gamma}{k} &=  \sum_{l=1}^K \sum_{i=1}^N \parpapb{Q}{}{\pi}{il}  \parpapb{\pi}{il} {\alpha}{ik} \parpapb{\alpha}{ik}{\gamma}{k},    \quad k=2,\ldots,K, \quad i=1,\ldots,N\\
\parpapb{Q}{}{\tau}{k} &=  \sum_{l=1}^K \sum_{i=1}^N \parpapb{Q}{}{\pi}{il} \parpapb{\pi}{il}{\alpha}{ik} \parpapb{\alpha}{ik}{\tau}{k},    \quad k=2,\ldots,K, \quad i=1,\ldots,N. \label{eq:dQdtau}
\end{align}
\end{subequations}
We have
\begin{equation}
\parpapb{Q}{}{\pi}{il} = \frac{y_{il}^{(q)}}{\pi_{il}}, \quad l=1,\ldots,K
\end{equation}
and
\begin{equation}
\label{eq:dpidalpha}
\parpapb{\pi}{il}{\alpha}{ik} = \begin{cases}
\displaystyle \frac{\sum_{q=1}^K \alpha_{iq}-\alpha_{il}}{\left(\sum_{q=1}^K \alpha_{iq} \right)^2} & \text{if } k=l\\
\displaystyle \frac{-\alpha_{il}}{\left(\sum_{q=1}^K \alpha_{iq} \right)^2} & \text{if } k\neq l
\end{cases}
\end{equation}
for all $k$ and $l$ in $\{1,\ldots,K\}$. 

Using the equality $\alpha_{ik}=\beta_k \Lambda(\gamma_k(t_i-\tau_k))$, we have
\begin{equation}
\label{eq:dalphadbeta}
\parpapb{\alpha}{ik}{\beta}{k}=\Lambda(\gamma_k(t_i-\tau_k))=\frac{\alpha_{ik}}{\beta_k},
\end{equation}
\begin{subequations}
\label{eq:dalphadgamma}
\begin{align}
\parpapb{\alpha}{ik}{\gamma}{k}&=\beta_k (t_i-\tau_k) \Lambda'(\gamma_k(t_i-\tau_k))\\
&=\alpha_{ik} (t_i-\tau_k) [1-\Lambda(\gamma_k(t_i-\tau_k)]\\
&=\alpha_{ik} (t_i-\tau_k) \left[1- \frac{\alpha_{ik}}{\beta_k} \right],
\end{align}
\end{subequations}
and
\begin{subequations}
\label{eq:dalphadtau}
\begin{align}
\parpapb{\alpha}{ik}{\tau}{k}&= - \gamma_k \, \beta_k \,  \Lambda'(\gamma_k(t_i-\tau_k))]\\
&= -\gamma_k \, \alpha_{ik}  \, [1-\Lambda(\gamma_k(t_i-\tau_k))]\\
&= - \gamma_k \,  \alpha_{ik} \left[1- \frac{\alpha_{ik}}{\beta_k} \right].
\end{align}
\end{subequations}

\begin{Rem} 
To impose the constraint $\gamma_1 = \ldots =\gamma_K$, we simply sum the derivatives w.r.t. $\gamma_k$:
\[
\parpapb{Q_r}{}{\gamma}{}= \sum_{k=1}^K  \parpapb{Q}{}{\gamma}{k},
\]
where $\parpapb{Q}{}{\gamma}{k}$ is computed using \eqref{eq:dQdgamma}.
\end{Rem}

When considering a prior on onsets, the derivatives w.r.t. $\tau_k$ \eqref{eq:dQdtau} become: 
\begin{equation}
\parpapb{Q_r}{}{\tau}{k} = \sum_{l=1}^K \sum_{i=1}^N \parpapb{Q}{}{\pi}{il} \parpapb{\pi}{il}{\alpha}{ik} \parpapb{\alpha}{ik}{\tau}{k} - 2 \lambda \biggl( \tau_k - \tau_k^{\textrm{prior}} \biggr),    \quad k=2,\ldots,K, \quad i=1,\ldots,N. \label{eq:Qr}
\end{equation}

Algorithm \ref{alg:gradient} summarises  the different steps of the gradient computation.

\begin{algorithm}[ht]
\caption{Gradient computation. \label{alg:gradient}}
\begin{algorithmic}
\REQUIRE $\beta_k$, $\gamma_k$, $\tau_k$, $k=2,\ldots,K$
\FOR{$i = 1$ \TO $N$}
\FOR{$k = 2$ \TO $K$}
\STATE Compute $\parpapb{\alpha}{ik}{\beta}{k}$ using \eqref{eq:dalphadbeta}
\STATE Compute $\parpapb{\alpha}{ik}{\gamma}{k}$ using \eqref{eq:dalphadgamma}
\STATE Compute $\parpapb{\alpha}{ik}{\tau}{k}$ using \eqref{eq:dalphadtau}
\FOR{$l = 1$ \TO $K$}
\STATE Compute $\parpapb{\pi}{il}{\alpha}{ik}$ using \eqref{eq:dpidalpha}
\ENDFOR
\ENDFOR
\ENDFOR
\FOR{$k = 2$ \TO $K$}
\STATE Compute $\parpapb{Q}{}{\beta}{k}$ using \eqref{eq:dQdbeta}
\STATE Compute $\parpapb{Q}{}{\gamma}{k}$ using \eqref{eq:dQdgamma}
\STATE Compute $\parpapb{Q}{ik}{\tau}{k}$ using \eqref{eq:dQdtau} or \eqref{eq:Qr}
\STATE Compute $\parpapb{Q}{}{b}{k}$ using \eqref{eq:dQdb}
\STATE Compute $\parpapb{Q}{}{g}{k}$ using \eqref{eq:dQdg}
\STATE Compute $\parpapb{Q}{}{\xi}{k}$ using \eqref{eq:dQdxi}
\ENDFOR
\ENSURE Gradient $\parpapb{Q}{}{\bomega}{}= \left\{\parpapb{Q}{}{b}{k}, \parpapb{Q}{}{g}{k}, \parpapb{Q}{}{\xi}{k}\right\}_{k=2}^K$
\end{algorithmic}
\end{algorithm}

\section{Summary of the hit detection and feature extraction steps}
\label{kharratetal}

{{
The AE signals were detected from the raw data stream using a method developed in \cite{kharrat2016signal}. This method is made of three steps: It starts by filtering the data stream, then applying a standard hit detection procedure, and finally feature extraction. The data were initially prefiltered using a high-pass filter of order $5$ with a band-pass frequency set to $10$ kHz and a band-pass ripple equal to $0.2$ dB in order to remove the DC component of the data stream. 

\paragraph{Step 1} The wavelet filtering step relies on wavelet denoising applied by frames of 250000 samples, practically shown in \cite{kharrat2016signal} to be a good compromise between computation time and quality of denoising with a better adaptation to noise and signals with varying properties. As advised in 
\cite{kharrat2016signal}, the wavelet was set to a Daubechies ``dB45'' made of $90$ coefficients, together with 14 levels of decomposition in order to detect the onset of AE signals. The soft Donoho-Johnstone universal threshold was applied on wavelet coefficients with a rescaling using a level-dependent estimation of level noise. A compensation of the group delay induced by the filtering was also performed. 

\paragraph{Step 2} After filtering, a hit detection is applied. The goal is to find the start and end samples for each AE signal remaining after filtering. For that, samples are treated gradually. When one sample becomes above a threshold on amplitude (in our case $1.2$ mV), samples are stored until one sample falls below the threshold. In this case, a counter is run. If the samples remain below the threshold during ``HDT'' microseconds (set to $1100 \,\mu s$), then the end of the signal is found. If not, the counter is reinitialised to 0. When HDT is reached, a second counter, ``HLT'' (set to $80 \,\mu s$), is run during which the detector is blind. It means that if the signal becomes above the threshold, it is not taken into account. This procedure is applied after filtering. However, the start and end are used to extract the AE signal in the raw data stream, not the filtered one (because the filtering alters the signals). This is one of the characteristics of the method.

\paragraph{Step 3} The features are finally extracted on each AE signal detected. Common features are described in 
\cite{pcimistras,sause12,kharrat2016signal,Kattis17}: Rise time, counts, PAC-energy, duration, amplitude, average frequency, RMS, average signal level, counts to peak, reverberation frequency, initiation frequency, signal strength, absolute energy, partial power in the intervals $[0, 20, 100, 200, 300, 400, 500, 600, 800, 1000]$ kHz, frequency centroid, peak frequency, weighted peak frequency. To this set of features were added the following ones: the Renyi number calculated from the scalogram \cite{gonzalez2000measuring} using a Morlet wavelet, as well as the frequency of the maximum of energy in the scalogram. 
For example features calculated in the frequency domain are well described in \cite{pcimistras,sause12}. Time-based features are described in \cite{pcimistras,kharrat2016signal,Kattis17}.
}}

\begin{figure}[hbtp]
     \centering
         \includegraphics[width=1\textwidth]{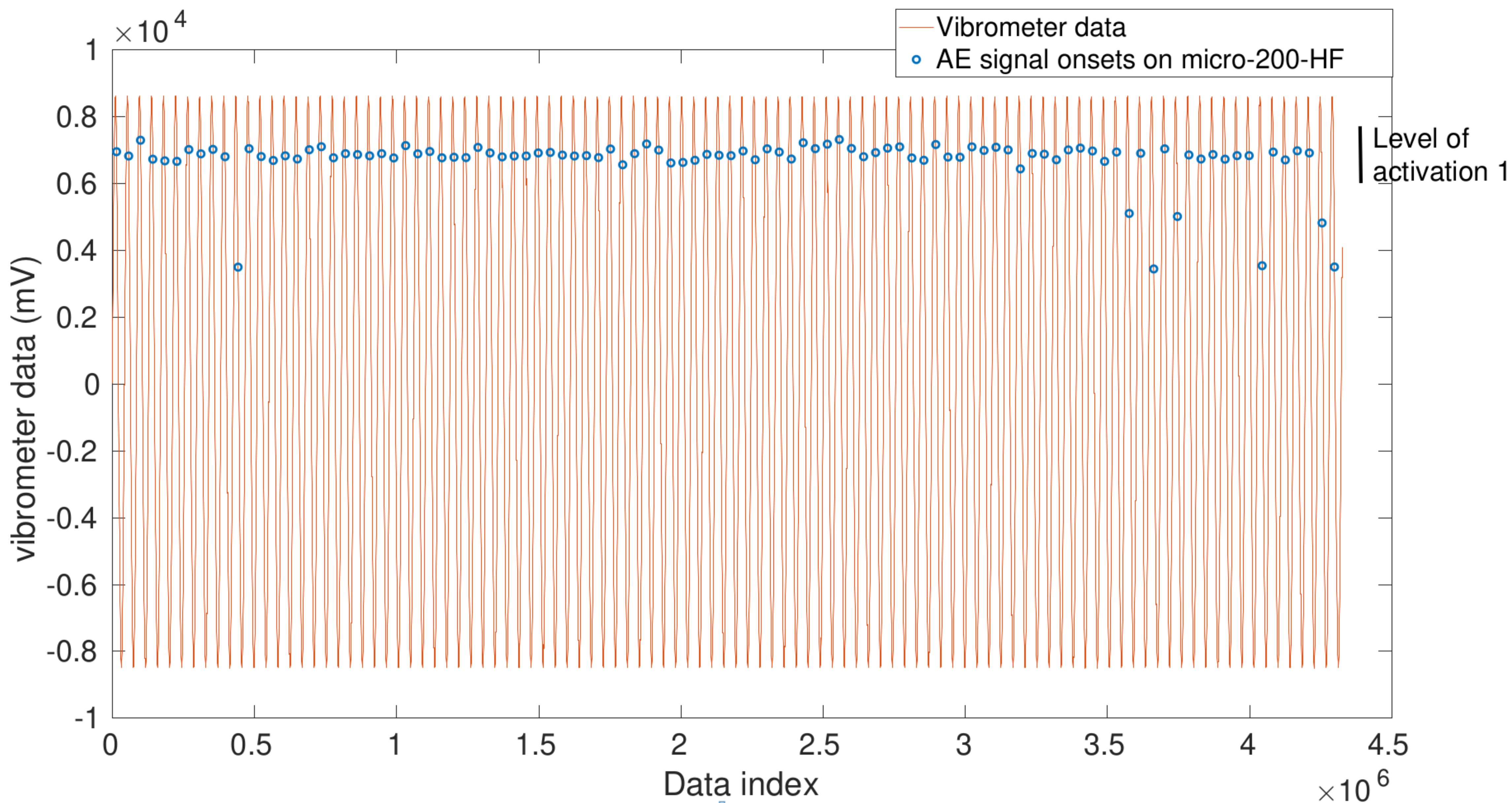}
        \caption{Position of onset of acoustic emission signals onto vibrometer data (for measurements B, 60 cNm, first file of the {{data set}}, using micro-200-HF sensor). The horizontal axis represents the sample index. \label{efzefzefzfzefze}}
\end{figure}

\paragraph{Illustration} {{After AE hit detection, the start time of each AE signal is}} superimposed onto vibrometer data for measurement B {{for a few cycles in Figure~\ref{efzefzefzfzefze}. We can observe that about 1-2 signals per cycle}} are found in this sample of about 1~s. {{This is confirmed in Table \ref{mlksapzkosa} that summarized the average over all campaigns and all files.}} We can also observe that the onsets are generally positioned on a similar displacement level (measured by the vibrometer) which {{means that the hit detection procedure finds signals mostly located at similar levels of displacement. This result demonstrates the reproducibility of the tests as well as the relevance of the hit detection.}} 

\clearpage


\begin{thebibliography}{10}
\expandafter\ifx\csname url\endcsname\relax
  \def\url#1{\texttt{#1}}\fi
\expandafter\ifx\csname urlprefix\endcsname\relax\def\urlprefix{URL }\fi
\expandafter\ifx\csname href\endcsname\relax
  \def\href#1#2{#2} \def\path#1{#1}\fi

\bibitem{ASTME1316}
{Subcommittee E07.92}, Standard terminology for nondestructive examinations,
  Standard, ASTM International, West Conshohocken, PA (2019).

\bibitem{ABS239}
Guidance notes on structural monitoring using acoustic emissions, Technical
  report, American Bureau of Shipping, City Plaza Drive, TX, USA (2016).

\bibitem{scruby1987introduction}
C.~Scruby, An introduction to acoustic emission, Journal of Physics E:
  Scientific Instruments 20~(8) (1987) 946.

\bibitem{FarrarBook}
C.~Farrar, K.~Worden, Structural Health Monitoring: A Machine Learning
  Perspective, John Wiley \& Sons, Ltd, 2013.

\bibitem{awerbuch2016applicability}
J.~Awerbuch, F.~Leone, D.~Ozevin, T.-M. Tan, On the applicability of acoustic
  emission to identify modes of damage in full-scale composite fuselage
  structures, Journal of Composite Materials 50~(4) (2016) 447--469.

\bibitem{bhuiyan2018toward}
M.~Y. Bhuiyan, J.~Bao, B.~Poddar, V.~Giurgiutiu, Toward identifying
  crack-length-related resonances in acoustic emission waveforms for structural
  health monitoring applications, Structural Health Monitoring 17~(3) (2018)
  577--585.

\bibitem{he2021overview}
Y.~He, M.~Li, Z.~Meng, S.~Chen, S.~Huang, Y.~Hu, X.~Zou, An overview of
  acoustic emission inspection and monitoring technology in the key components
  of renewable energy systems, Mechanical Systems and Signal Processing 148
  (2021) 107146.

\bibitem{alshorman2021sounds}
O.~AlShorman, F.~Alkahatni, M.~Masadeh, M.~Irfan, A.~Glowacz, F.~Althobiani,
  J.~Kozik, W.~Glowacz, Sounds and acoustic emission-based early fault
  diagnosis of induction motor: A review study, Advances in Mechanical
  Engineering 13~(2) (2021) 1687814021996915.

\bibitem{kurz2005strategies}
J.~H. Kurz, C.~U. Grosse, H.-W. Reinhardt, Strategies for reliable automatic
  onset time picking of acoustic emissions and of ultrasound signals in
  concrete, Ultrasonics 43~(7) (2005) 538--546.

\bibitem{Pomponi2015110}
E.~Pomponi, A.~Vinogradov, A.~Danyuk, Wavelet based approach to signal activity
  detection and phase picking: Application to acoustic emission, Signal
  Processing 115 (2015) 110 -- 119.

\bibitem{bianchi2015wavelet}
D.~Bianchi, E.~Mayrhofer, M.~Gr{\"o}schl, G.~Betz, A.~Vernes, Wavelet packet
  transform for detection of single events in acoustic emission signals,
  Mechanical Systems and Signal Processing 64 (2015) 441--451.

\bibitem{WARRENLIAO201074}
T.~W. Liao, Feature extraction and selection from acoustic emission signals
  with an application in grinding wheel condition monitoring, Engineering
  Applications of Artificial Intelligence 23~(1) (2010) 74 -- 84.

\bibitem{kharrat2016signal}
M.~Kharrat, E.~Ramasso, V.~Placet, M.~Boubakar, A signal processing approach
  for enhanced acoustic emission data analysis in high activity systems:
  Application to organic matrix composites, Mechanical Systems and Signal
  Processing 70 (2016) 1038--1055.

\bibitem{madarshahian2019acoustic}
R.~Madarshahian, P.~Ziehl, J.~M. Caicedo, Acoustic emission bayesian source
  location: onset time challenge, Mechanical Systems and Signal Processing 123
  (2019) 483--495.

\bibitem{RAMASSO2020103478}
E.~Ramasso, P.~Butaud, T.~Jeannin, F.~Sarasini, V.~Placet, N.~Godin,
  J.~Tirillò, X.~Gabrion, Learning the representation of raw acoustic emission
  signals by direct generative modelling and its use in chronology-based
  clusters identification, Engineering Applications of Artificial Intelligence
  90 (2020) 103478.

\bibitem{Kattis17}
S.~Kattis, Noesis: Advanced data analysis, pattern recognition \& neural
  networks software for acoustic emission applications, in: Kolloquium
  Schallemission, Statusberichte zur Entwicklung und Anwendung der
  Schallemissionsanalyse, Vol.~12, Fulda, 2017, pp. 1--8.

\bibitem{manson2001visualisation}
G.~Manson, K.~Worden, K.~Holford, R.~Pullin, Visualisation and dimension
  reduction of acoustic emission data for damage detection, Journal of
  Intelligent Material Systems and Structures 12~(8) (2001) 529--536.

\bibitem{li2012feature}
M.~Li, J.~H. Yang, Feature selection of acoustic emission signal for the
  slow-speed and heavy-load equipment, in: Applied Mechanics and Materials,
  Vol. 110, Trans Tech Publ, 2012, pp. 3199--3203.

\bibitem{doan2015unsupervised}
D.~Doan, E.~Ramasso, V.~Placet, S.~Zhang, L.~Boubakar, N.~Zerhouni, An
  unsupervised pattern recognition approach for {AE} data originating from
  fatigue tests on polymer--composite materials, Mechanical Systems and Signal
  Processing 64 (2015) 465--478.

\bibitem{sause16}
M.~G. Sause, In Situ Monitoring of Fiber-Reinforced Composites: Theory, basic
  concepts, methods, and applications, springer series in materials science
  Edition, Vol. 242, Springer Int. Publishing, 2016.

\bibitem{ramasso2015unsupervised}
E.~Ramasso, V.~Placet, M.~Boubakar, Unsupervised consensus clustering of
  acoustic emission time-series for robust damage sequence estimation in
  composites, IEEE Trans. on Instr. and Meas. 64~(12) (2015) 3297--3307.

\bibitem{MARTINDELCAMPO2017187}
S.~Martin-Del-Campo, F.~Sandin, Online feature learning for condition
  monitoring of rotating machinery, Engineering Applications of Artificial
  Intelligence 64 (2017) 187 -- 196.

\bibitem{FUENTES2020776}
R.~Fuentes, R.~Dwyer-Joyce, M.~Marshall, J.~Wheals, E.~Cross, Detection of
  sub-surface damage in wind turbine bearings using acoustic emissions and
  probabilistic modelling, Renewable Energy 147 (2020) 776 -- 797.

\bibitem{wang2019tool}
C.~Wang, Z.~Bao, P.~Zhang, W.~Ming, M.~Chen, Tool wear evaluation under minimum
  quantity lubrication by clustering energy of acoustic emission burst signals,
  Measurement 138 (2019) 256--265.

\bibitem{chelliah2019optimization}
S.~K. Chelliah, P.~Parameswaran, S.~Ramasamy, A.~Vellayaraj, S.~Subramanian,
  Optimization of acoustic emission parameters to discriminate failure modes in
  glass--epoxy composite laminates using pattern recognition, Structural Health
  Monitoring 18~(4) (2019) 1253--1267.

\bibitem{zhou2018cluster}
W.~Zhou, W.-z. Zhao, Y.-n. Zhang, Z.-j. Ding, Cluster analysis of acoustic
  emission signals and deformation measurement for delaminated glass fiber
  epoxy composites, Composite Structures 195 (2018) 349--358.

\bibitem{MacQueen1967}
J.~MacQueen, Some methods for classification and analysis of multivariate
  observations, in: Proc. of the Fifth Berkeley Symposium on Math., Stat. and
  Prob., Vol.~1, 1967, pp. 281--296.

\bibitem{chai2017acoustic}
M.~Chai, J.~Zhang, Z.~Zhang, Q.~Duan, G.~Cheng, Acoustic emission studies for
  characterization of fatigue crack growth in 316ln stainless steel and welds,
  Applied Acoustics 126 (2017) 101--113.

\bibitem{dunn1973fuzzy}
J.~C. Dunn, A fuzzy relative of the isodata process and its use in detecting
  compact well-separated clusters, Journal of Cybernetics 3 (1973) 32--57.

\bibitem{1198989}
S.~N. {Omkar}, S.~{Suresh}, T.~R. {Raghavendra}, V.~{Mani}, Acoustic emission
  signal classification using fuzzy c-means clustering, in: Proceedings of the
  9th International Conference on Neural Information Processing, 2002. ICONIP
  '02., Vol.~4, 2002, pp. 1827--1831 vol.4.

\bibitem{gustafson1979fuzzy}
D.~E. Gustafson, W.~C. Kessel, Fuzzy clustering with a fuzzy covariance matrix,
  in: 1978 IEEE conference on decision and control including the 17th symposium
  on adaptive processes, IEEE, 1979, pp. 761--766.

\bibitem{mclachlan1988mixture}
G.~J. McLachlan, K.~E. Basford, Mixture models. Inference and applications to
  clustering, Vol.~84, Marcel Dekker, Statistics: Textbooks and Monographs, New
  York, United States, 1988.

\bibitem{sawan2015unsupervised}
H.~A. Sawan, M.~E. Walter, B.~Marquette, Unsupervised learning for
  classification of acoustic emission events from tensile and bending
  experiments with open-hole carbon fiber composite samples, Composites Science
  and Technology 107 (2015) 89--97.

\bibitem{kaminski2015fatigue}
M.~Kaminski, F.~Laurin, J.~Maire, C.~Rakotoarisoa, E.~H{\'e}mon, Fatigue damage
  modeling of composite structures: the onera viewpoint, AerospaceLab 6~(9)
  (2015) 1--12.

\bibitem{saxena2011accelerated}
A.~Saxena, K.~Goebel, C.~C. Larrosa, V.~Janapati, S.~Roy, F.-K. Chang,
  Accelerated aging experiments for prognostics of damage growth in composite
  materials, Tech. rep., NASA, Moffett Field CA Ames Research (2011).

\bibitem{PATTERNRECONliao2005clustering}
T.~W. Liao, Clustering of time series data - a survey, Pattern recognition
  38~(11) (2005) 1857--1874.

\bibitem{fu2011review}
T.-C. Fu, A review on time series data mining, Engineering Applications of
  Artificial Intelligence 24~(1) (2011) 164--181.

\bibitem{belhadi2020space}
A.~Belhadi, Y.~Djenouri, K.~N{\o}rv{\aa}g, H.~Ramampiaro, F.~Masseglia,
  J.~C.-W. Lin, Space--time series clustering: Algorithms, taxonomy, and case
  study on urban smart cities, Engineering Applications of Artificial
  Intelligence 95 (2020) 103857.

\bibitem{moller2003fuzzy}
C.~S. M{\"o}ller-Levet, F.~Klawonn, K.-H. Cho, O.~Wolkenhauer, Fuzzy clustering
  of short time-series and unevenly distributed sampling points, in:
  International symposium on intelligent data analysis, Springer, 2003, pp.
  330--340.

\bibitem{sause12}
M.~Sause, A.~Gribov, A.~Unwin, S.~Horn, Pattern recognition approach to
  identify natural clusters of acoustic emission signals, Pattern Reco. Lett.
  33 (2012) 17--23.

\bibitem{Godin18}
N.~Godin, P.~Reynaud, G.~Fantozzi, Acoustic emission and durability of
  composites materials, ISTE-Wiley editions, 2018.

\bibitem{monti2016mechanical}
A.~Monti, A.~El~Mahi, Z.~Jendli, L.~Guillaumat, Mechanical behaviour and damage
  mechanisms analysis of a flax-fibre reinforced composite by acoustic
  emission, Composites Part A: Applied Science and Manufacturing 90 (2016)
  100--110.

\bibitem{SHIRAIWA20202791}
T.~Shiraiwa, K.~Ishikawa, M.~Enoki, I.~Shinozaki, S.~Kanazawa, Acoustic
  emission analysis using bayesian model selection for damage characterization
  in ceramic matrix composites, Journal of the European Ceramic Society 40~(8)
  (2020) 2791 -- 2800.

\bibitem{carmichael1968finding}
J.~Carmichael, R.~Julius, Finding natural clusters, Systematic Biology 17~(2)
  (1968) 144--150.

\bibitem{placet2013online}
V.~Placet, E.~Ramasso, L.~Boubakar, N.~Zerhouni, Online segmentation of
  acoustic emission data streams for detection of damages in composites
  structures in unconstrained environments, in: 11th Int. Conf. on Structural
  Safety \& Reliability, 2013, pp. 1--8.

\bibitem{hohl2018computationally}
A.~Hohl, A.~D~Griffith, M.~C. Eppes, E.~Delmelle, Computationally enabled 4d
  visualizations facilitate the detection of rock fracture patterns from
  acoustic emissions, Rock Mechanics and Rock Engineering (2018) 1--14.

\bibitem{rastegaev2018using}
I.~Rastegaev, D.~Merson, A.~Danyuk, M.~Afanasyev, A.~Vinogradov, Using acoustic
  emission signal categorization for reconstruction of wear development
  timeline in tribosystems: Case studies and application examples, Wear (2018)
  83--92.

\bibitem{8529965}
L.~{Li}, Y.~{Pu}, J.~{Chen}, Maximum likelihood parameter estimation for armax
  models based on stochastic gradient algorithm, in: 2018 10th International
  Conference on Modelling, Identification and Control (ICMIC), 2018, pp. 1--6.

\bibitem{Vinogradov20}
I.~Rastegaev, D.~Merson, I.~Rastegaeva, A.~Vinogradov, A time-frequency based
  approach for acoustic emission assessment of sliding wear, Lubricants 52~(7)
  (2020) 1--24.

\bibitem{NehaPhD}
N.~Chandarana, Combining passive and active methods for damage mode diagnosis
  in tubular composites, Ph.D. thesis, Manchester University, Faculty of
  Science and Engineering, Departement of Materials (11/18/2019).

\bibitem{orionaedata}
B.~Verdin, G.~Chevallier, E.~Ramasso, {ORION-AE}: Multisensor acoustic emission
  datasets reflecting supervised untightening of bolts in a jointed vibrating
  structure (Harvard Dataverse, 2021).
\newblock \href {http://dx.doi.org/10.7910/DVN/FBRDU0}
  {\path{doi:10.7910/DVN/FBRDU0}}.

\bibitem{FRIGIERI2016230}
E.~P. Frigieri, P.~H. Campos, A.~P. Paiva, P.~P. Balestrassi, Joao, A
  mel-frequency cepstral coefficient-based approach for surface roughness
  diagnosis in hard turning using acoustic signals and gaussian mixture models,
  Applied Acoustics 113 (2016) 230 -- 237.

\bibitem{PREM201728}
P.~R. Prem, A.~R. Murthy, Acoustic emission monitoring of reinforced concrete
  beams subjected to four-point-bending, Applied Acoustics 117 (2017) 28 -- 38.

\bibitem{SAGAR2018647}
R.~V. Sagar, J.~Srivastava, R.~Singh, A probabilistic analysis of acoustic
  emission events and associated energy release during formation of shear and
  tensile cracks in cementitious materials under uniaxial compression, Journal
  of Building Engineering 20 (2018) 647 -- 662.

\bibitem{DAS201942}
A.~K. Das, D.~Suthar, C.~K. Leung, Machine learning based crack mode
  classification from unlabeled acoustic emission waveform features, Cement and
  Concrete Research 121 (2019) 42 -- 57.

\bibitem{dempster77}
A.~Dempster, N.~Laird, D.~Rubin, Maximum likelihood from incomplete data via
  the {EM} algorithm, Journal of the Royal Statistical Society 39~(1) (1977)
  1--38.

\bibitem{mclachlan08}
G.~J. McLachlan, T.~Krishnan, The {EM} Algorithm and Extensions, 2nd Edition,
  Wiley, New York, 2008.

\bibitem{ChatterjeeGEM}
S.~Chatterjee, O.~Romero, S.~Pequito, Analysis of a generalised
  expectation–maximisation algorithm for gaussian mixture models: a control
  systems perspective, International Journal of Control 0~(0) (2021) 1--9.

\bibitem{biernacki1999choosing}
C.~Biernacki, G.~Govaert, Choosing models in model-based clustering and
  discriminant analysis, Journal of Statistical Computation and Simulation
  64~(1) (1999) 49--71.

\bibitem{baudry2010combining}
J.-P. Baudry, A.~E. Raftery, G.~Celeux, K.~Lo, R.~Gottardo, Combining mixture
  components for clustering, Journal of computational and graphical statistics
  19~(2) (2010) 332--353.

\bibitem{ORIONdata}
E.~Ramasso, B.~Verdin, G.~Chevallier, Monitoring a bolted vibrating structure
  using multiple acoustic emission sensors: A benchmark, MDPI DATA 7 (2022)
  31--45.

\bibitem{zhang2019continuous}
Z.~Zhang, Y.~Xiao, Z.~Su, Y.~Pan, Continuous monitoring of tightening condition
  of single-lap bolted composite joints using intrinsic mode functions of
  acoustic emission signals: a proof-of-concept study, Structural Health
  Monitoring 18~(4) (2019) 1219--1234.

\bibitem{pcimistras}
M.~of~MISTRAS Holdings~Group, Pci-2 based ae system user's manual rev. 3
  (2007).

\bibitem{gonzalez2000measuring}
S.~Gonzalez~Andino, R.~Grave~de Peralta~Menendez, G.~Thut, L.~Spinelli,
  O.~Blanke, C.~Michel, T.~Landis, Measuring the complexity of time series: an
  application to neurophysiological signals, Human brain mapping 11~(1) (2000)
  46--57.

\bibitem{ari}
N.~Vinh, J.~Epps, J.~Bailey, Information theoretic measures for clusterings
  comparison: Is a correction for chance necessary?, in: Proc. of the 26th
  Annual Int. Conference on Machine Learning, ACM, New York, NY, USA, 2009, pp.
  1073--1080.

\end{thebibliography}

\end{document}